\documentclass{article}

\usepackage{microtype}
\usepackage{graphicx}
\usepackage{xcolor}
\usepackage{booktabs} 
\usepackage{authblk}
\usepackage{natbib}
\usepackage{graphicx}
\usepackage{caption}
\usepackage{subcaption}
\usepackage{multirow}
\usepackage[a4paper, total={6in, 8in}]{geometry}
\usepackage[ruled,vlined]{algorithm2e}
\usepackage{hyperref}

\usepackage{amsmath,amsfonts,bm}

\def\eqref#1{equation~\ref{#1}}

\def\1{\bm{1}}

\def\rvb{{\mathbf{b}}}

\def\rvf{{\mathbf{f}}}
\def\rvg{{\mathbf{g}}}

\def\rvn{{\mathbf{n}}}
\def\rvo{{\mathbf{o}}}
\def\rvp{{\mathbf{p}}}
\def\rvq{{\mathbf{q}}}

\def\rvx{{\mathbf{x}}}
\def\rvy{{\mathbf{y}}}
\def\rvz{{\mathbf{z}}}

\def\rmA{{\mathbf{A}}}
\def\rmB{{\mathbf{B}}}

\def\rmD{{\mathbf{D}}}

\def\rmG{{\mathbf{G}}}

\def\rmM{{\mathbf{M}}}
\def\rmN{{\mathbf{N}}}

\def\rmQ{{\mathbf{Q}}}

\def\rmW{{\mathbf{W}}}
\def\rmX{{\mathbf{X}}}
\def\rmY{{\mathbf{Y}}}

\DeclareMathAlphabet{\mathsfit}{\encodingdefault}{\sfdefault}{m}{sl}
\SetMathAlphabet{\mathsfit}{bold}{\encodingdefault}{\sfdefault}{bx}{n}

\newcommand{\softmax}{\mathrm{softmax}}

\newcommand{\Var}{\mathrm{Var}}

\makeatletter
\def\@opargbegintheorem#1#2#3{\trivlist
   \item[]{\bfseries #1\ #2\ (#3)} \itshape}
\makeatother

\newcommand{\norm}[1]{\|#1\|}
\newcommand{\highlight}[1]{\textbf{\textit{#1}}}
\newcommand{\expectation}[1]{\mathbb{E}\left({#1}\right)}
\newtheorem{theorem}{Theorem}   
\newtheorem{lemma}{Lemma}        
\newtheorem{assumption}{Assumption}  
\newtheorem{corollary}{Corollary}       
\newtheorem{problem}{Problem}                 
  
\newtheorem{remark}{Remark} 
\newtheorem{definition}{Definition}
\usepackage{graphicx}

\begin{document}

\author[1]{Boyang Liu}
\author[1*]{Mengying Sun}
\author[1*]{Ding Wang}
\author[1*]{Pang-Ning Tan}
\author[1*]{Jiayu Zhou}
\affil[1]{Department of Computer Science, Michigan State University}
\date{}
\title{Learning Deep Neural Networks under Agnostic Corrupted Supervision}

\maketitle

\vskip 0.3in

\begin{abstract}
Training deep neural models in the presence of corrupted supervision is challenging as the corrupted data points may significantly impact the generalization performance.
To alleviate this problem, we present an efficient robust algorithm that achieves strong guarantees without any assumption on the type of corruption, and provides a unified framework for both classification and regression problems. Unlike many existing approaches that quantify the quality of the data points (e.g., based on their individual loss values), and filter them accordingly, 
the proposed algorithm focuses on controlling the collective impact of data points on the average gradient. 
Even when a corrupted data point failed to be excluded by our algorithm, the data point will have very limited impact on the overall loss, as compared with state-of-the-art filtering methods based on loss values. 
Extensive experiments on multiple benchmark datasets have demonstrated the robustness of our algorithm under different types of corruptions.
\end{abstract}

\section{Introduction}
Corrupted supervision is a common issue in real-world learning tasks, where the learning targets are potentially noisy due to errors in the data collection or labeling process.
Such corruptions can have severe consequences especially in deep learning models, whose large degree-of-freedom makes them easier to memorize the corrupted examples, and thus, susceptible to overfitting~\citep{zhang2016understanding}. 

There have been extensive efforts to achieve robustness against corrupted supervision. 
A natural approach to deal with corrupted supervision in deep neural networks (DNNs) is to reduce the model exposure to corrupted data points during training. By detecting and filtering (or re-weighting) the possible corrupted samples, the learning is expected to deliver a model that is similar to the one trained on clean data (without corruption) \citep{kumar2010self, han2018co, zhengerror}. 
There are various criteria designed to identify the corrupted data points in training. For example, 
\citet{kumar2010self, han2018co, jiang2018mentornet} leveraged the loss function values of the data points; 
\citet{zhengerror} 
considered prediction uncertainty for filtering data; \citet{malach2017decoupling} used the disagreement between two deep networks;  while
\citet{reed2014training} utilized the prediction consistency of neighboring iterations. 
The success of these methods highly depends on the effectiveness of the detection criteria in correctly identifying the corrupted data points. 
Since the true corrupted points remain unknown throughout the learning process, such  ``unsupervised" methods may not be effective, either lacking in theoretical guarantees of robustness~\citep{han2018co, reed2014training, malach2017decoupling, li2017learning} or providing guarantees only under the assumption 
that prior knowledge is available about the type of corruption present~\citep{zhengerror, shah2020choosing, patrini2017making,  yi2019probabilistic}. 
Most existing theoretical guarantees under agnostic corruption during optimization are focused on convex losses~\citep{prasad2018robust} or linear models~\citep{bhatia2015robust, bhatia2017consistent}, and thus cannot be directly extended to DNNs. \citep{diakonikolas2019sever} developed a generalized non-convex optimization algorithm against agnostic corruptions. However, it is not optimized for the label/supervision corruption and has a high space complexity, which may incur prohibitive costs when applied to typical DNNs with a large amount of parameters. 
Furthermore, 
many existing approaches 
are exclusively designed for classification problems (e.g., \cite{malach2017decoupling, reed2014training, menon2019can, zhengerror}); 
extending them to solve regression problems is not that straightforward.

To tackle these challenges, this paper presents a unified optimization framework with robustness guarantees, without any assumptions on how supervisions are corrupted, and is applicable to both classification and regression problems. Instead of designing a criterion for accurate detection of  corrupted samples, we focus on limiting the collective impact of corrupted samples during the learning process through \emph{robust mean estimation} of the gradients. Specifically, if our estimated average gradient is close to the expected gradient from the clean data during the learning iterations, then the final model will be close to the model trained on clean data. As such, a corrupted data point can still be used during the training as long as it does not considerably alter the average gradient. This observation has remarkably impact  our algorithm design: instead of explicitly quantifying (and identifying) individual corrupted data points, which is a hard problem in itself, we are now dealing with an easier task, i.e., eliminating training data points that significantly distort the mean gradient estimation. 
One immediate consequence of this design is that, even when a corrupted data point failed to be excluded by the proposed algorithm, the data point will likely have very limited impact on the overall gradient, as compared with state-of-the-art filtering data points based on loss values. 
Compared to state-of-the-art robust optimization methods~\citep{prasad2018robust, diakonikolas2019sever} that require the more expensive SVD computation on the gradient matrix, we fully utilize the gradient structure when the corruptions are exclusively on the supervision to make our algorithm applicable to DNNs. Moreover, when only supervision are corrupted, we improve the error bound from $\mathcal{O}(\sqrt{\epsilon})$ to $\mathcal{O}(\epsilon)$, where $\epsilon$ is the corruption rate.
We perform experiments on both regression and classification with corrupted supervision on multiple benchmark datasets. The results show that the proposed method outperforms state-of-the-art.

\section{Background}
Learning from corrupted data \citep{huber1992robust} has attracted considerable attention in the machine learning community \citep{natarajan2013learning}. Many recent studies have investigated robustness of classification tasks with noisy labels. For example, \citet{kumar2010self} proposed a self-paced learning (SPL) approach, which assigns higher weights to examples with smaller loss. A similar idea was used in curriculum learning \citep{bengio2009curriculum}, in which the model learns the easy concept before the harder ones. Alternative methods inspired by SPL include learning the data weights \citep{jiang2018mentornet} and collaborative learning \citep{han2018co, yu2019does}. 
Label correction \citep{patrini2017making, li2017learning, yi2019probabilistic} is another approach, which attempts to revise the original labels of the data 
to recover clean labels from corrupted ones. However, since we do not have access to which data points are corrupted, it is harder to obtain provable guarantees for label correction without strong assumptions about the corruption type.

Accurate estimation of the gradient is a key step for successful optimization. The relationship between gradient estimation and its final convergence has been widely studied in the optimization community. Since computing an approximated (and potentially biased) gradient is often more efficient than computing the exact gradient, many studies used approximated gradients to optimize their models and showed that they suffer from the biased estimation problem if 
there is no assumption on the gradient estimation \citep{d2008smooth,schmidt2011convergence,bernstein2018signsgd, hu2020biased, ajalloeian2020analysis}. 

A closely related topic is robust estimation of the mean.
Given corrupted data, robust mean estimation aims at generating an estimated mean $\hat{\mathbf{\mu}}$ such that the difference between the estimated mean on corrupted data and the mean of clean data  $\norm{\hat{\mathbf{\mu}} - \mathbf{\mu}}_2$  is minimized. The median or trimmed mean have been shown to be optimal statistics for mean estimation in one-dimensional data \citep{huber1992robust}. However, robustness in high dimension is more challenging since applying the coordinate-wise optimal robust estimator would lead to an error factor $\mathcal{O}(\sqrt{d})$ that scales with dimensionality of the data. Although classical methods such as Tukey median \citep{tukey1975mathematics} have successfully designed algorithms to eliminate the  $\mathcal{O}(\sqrt{d})$ error, the algorithms cannot run in polynomial-time. More recently, \citet{diakonikolas2016robust, lai2016agnostic} successfully designed polynomial-time algorithms with dimension-free error bounds. 
The results have been widely applied to improve algorithmic efficiency in various scenarios~\citep{dong2019quantum, cheng2020high}.

Robust optimization is designed to improve algorithm robustness in the presence of corrupted data. 
Most existing efforts have focused on linear regression and its variants~\citep{bhatia2015robust,bhatia2017consistent, shen2019learning} or convex problems~\citep{prasad2018robust}. Thus, their results cannot be directly generalized to DNNs. Although \citet{diakonikolas2019sever} presented a generalized non-convex optimization method with an agnostic corruption guarantee, the space complexity of the algorithm is high, and cannot be applied to DNNs due to large parameter sizes. We will discuss \citep{diakonikolas2019sever} in more details in the next section.

\section{Methodology}
Before introducing our algorithm, we first present our corrupted supervision setting. To characterize agnostic corruptions, we 
assume there is an \emph{adversary} that tries to corrupt the supervision of clean data. There is no restriction on how the adversary corrupts the supervision, which can either be randomly permuting the target, or in a way that maximizes negative impact (i.e., lower performance) on the model. The adversary can choose up to $\epsilon$ fraction of the clean target $\rmD_y \in \mathbb{R}^{n \times q}$ and alter the selected rows of $\rmD_y$ to arbitrary valid numbers, generating $\rmD_y^{\epsilon} \in \mathbb{R}^{n \times q}$. The adversary then returns the corrupted dataset $\rmD_x$, $\rmD_y^{\epsilon}$ to our learning algorithm $\mathcal{A}$. The adversary can have full knowledge of the data or even the learning algorithm $\mathcal{A}$. The only constraint on the adversary is the corruption rate, $\epsilon$. A key question is: \emph{Given a dataset $\rmD_x \in \mathbb{R}^{n \times p}$, $\rmD_y^{\epsilon} \in \mathbb{R}^{n \times q}$, with $\epsilon$-fraction of corrupted supervision, and a learning objective $\phi: \mathbb{R}^p \times \mathbb{R}^q \times \mathbb{R}^d  \rightarrow \mathbb{R}$ parameterized by $\theta  \in \mathbb{R}^d$, can we output the parameters $\mathbf{\theta}$ such that $\norm{\nabla_{\mathbf{\mathbf{\theta}}} \phi(\mathbf{\theta};\rmD_x, \rmD_y)}$ is minimized}?

When $\epsilon = 0$, $\rmD_y^{\epsilon} = \rmD_y$ and the learning is performed on  clean data. The stochastic gradient descent algorithm may converge to a stationary point where $\norm{\nabla_{\mathbf{\mathbf{\theta}}} \phi(\mathbf{\theta};\rmD_x, \rmD_y)} = 0$. 
However, this is no longer the case when the supervision is corrupted as above 
due to the impact of the corrupted data on $\mathbf{\theta}$. 
We thus want an efficient algorithm to find a model $\mathbf{\theta}$ that minimizes $\norm{\nabla_{\mathbf{\mathbf{\theta}}} \phi(\mathbf{\theta};\rmD_x, \rmD_y)}$. A robust model $\mathbf{\theta}$ should have a small value of $\norm{\nabla_{\mathbf{\mathbf{\theta}}} \phi(\mathbf{\theta};\rmD_x, \rmD_y)}$, 
and we hypothesize that a smaller $\norm{\nabla_{\mathbf{\mathbf{\theta}}} \phi(\mathbf{\theta};\rmD_x, \rmD_y)}$ leads to better generalization.

\subsection{Stochastic Gradient Descent with Biased Gradient}
A direct consequence of corrupted supervision is biased gradient estimation. 
In this section, we will first analyze how such biased gradient estimation affects the robustness of learning.
The classical analysis of stochastic gradient descent (SGD) requires access to the stochastic gradient oracle, which is an unbiased estimation of the true gradient.
However, corrupted supervision leads to corrupted gradients, which makes it   difficult to get unbiased gradient estimation without assumptions on how the gradients are corrupted.

The convergence of biased gradient has been studied via a series of previous works~\citep{schmidt2011convergence, bernstein2018signsgd, hu2020biased, ajalloeian2020analysis, scaman2020robustness}. We show a similar theorem below for the sake of completeness. Before We gave the theorem of how biased gradient affect the final convergence of SGD. We introduce several assumptions and definition first:
\begin{assumption}[L-smoothness]
	\label{asm:L-smooth}
	The function $\phi$: $\mathbb{R}^d \rightarrow \mathbb{R}$ is differentiable and there exists a constant $L > 0$ such that for all $\mathbf{\theta}_1, \mathbf{\theta}_2 \in \mathbb{R}^d$, we have $\phi(\mathbf{\theta}_2) \leq \phi(\mathbf{\theta}_1)+\langle\nabla \phi(\mathbf{\theta}_1), \mathbf{\theta}_2-\mathbf{\theta}_1\rangle+\frac{L}{2}\|\mathbf{\theta}_2-\mathbf{\theta}_1\|^{2}$
\end{assumption}
\begin{definition}[Biased gradient oracle]
	A map $\rvg: \mathbb{R}^d \times \mathcal{D} \rightarrow \mathbb{R}^d$, such that $\mathbf{g}(\mathbf{\theta}, \xi)=\nabla \phi(\mathbf{\theta})+\mathbf{b}(\mathbf{\theta}, \xi)+\mathbf{n}(\mathbf{\theta}, \xi)$ for a bias $\rvb: \mathbb{R}^d  \rightarrow \mathbb{R}^d$ and zero-mean noise  $\rvn: \mathbb{R}^d \times \mathcal{D} \rightarrow \mathbb{R}^d$, that is $\mathbb{E}_{\xi}\left(\rvn(\mathbf{\theta}, \xi)\right) = 0$.
\end{definition}
Compared to standard stochastic gradient oracle, the above definition introduces the bias term $\rvb$. In noisy-label settings, the $\rvb$ is generated by the data with corrupted labels. 

\begin{assumption}[$\sigma$-Bounded noise]
	\label{asm:boundednoise}
	There exists constants $\sigma > 0$, such that
	$\mathbb{E}_{\xi}\|\mathbf{n}(\mathbf{\theta}, \xi)\|^{2} \leq \sigma, \quad \forall \mathbf{\theta} \in \mathbb{R}^{d}$
\end{assumption}

\begin{assumption}[$\zeta$-Bounded bias]
	\label{asm:boundedbias}
	There exists constants $\zeta > 0$, such that for any $\xi$, we have
	$\|\mathbf{b}(\mathbf{\theta}, \xi)\|^{2} \leq \zeta^2, \quad \forall \mathbf{\theta} \in \mathbb{R}^{d}$
\end{assumption}

For simplicity, assume the learning rate is constant $\gamma$, then in every iteration, the biased SGD performs update $\mathbf{\theta}_{t+1} \leftarrow \mathbf{\theta}_{t} - \gamma_{t}\rvg(\mathbf{\theta}_{t}, \xi)$.  Then the following theorem showed the gradient norm convergence with biased SGD.

\begin{theorem}[Convergence of Biased SGD(formal)] \label{theo:convergence} Under assumptions \ref{asm:L-smooth}, \ref{asm:boundednoise}, \ref{asm:boundedbias}, define $F = \phi(\mathbf{\theta}_0)-\phi^*$and step size $\gamma = \min\left\{ \dfrac{1}{L}, (\sqrt{\dfrac{LF}{\sigma T})}\right\}$, denote the desired accuracy as $k$, then 
	\begin{align*}
	T=\mathcal{O}\left(\frac{1}{k}+\frac{\sigma^{2}}{k^{2}}\right) 
	\end{align*}
	iterations are sufficient to obtain $\min_{t \in [T]}\expectation{\norm{\nabla \phi(\mathbf{\theta}_t)}^2} = \mathcal{O}(k + \zeta^2)$. 
\end{theorem}
\begin{remark}
	Let $k = \zeta^2$, $T=\mathcal{O}\left(\frac{1}{\zeta^2}+\frac{\sigma^{2}}{
		\zeta^{4}}\right)$ iterations is 
		sufficient to get \\
		$\min_{t \in [T]}\expectation{\norm{\nabla \phi(\mathbf{\theta}_t)}^2} = \mathcal{O}(\zeta^2)$, and performing more iterations does not improve the accuracy in terms of convergence.
\end{remark}

The difference between the above theorem and the typical convergence theorem for SGD is that we are using a biased gradient estimation. 
According to Theorem~\ref{theo:convergence},  robust estimation of the gradient $\rvg$ is the key to ensure a robust model that converges to the clean solution. We also assume the loss function has the form of $\mathcal{L}(\rvy, \hat{\rvy})$, in which many commonly used loss functions belong to this category.

\subsection{Robust Gradient Estimation for General Data Corruption}
Before discussing the corrupted supervision setting, we first review the general corruption setting, where the corruptions may be present in both the supervision and input features. 
A na\"ive approach is 
to apply a robust coordinate-wise gradient estimation approach such as 
coordinate-wise median for gradient estimation. However, by using the coordinate-wise robust estimator, the L2 norm of the difference between the estimated and ground-truth gradients contains a factor of $\mathcal{O}(\sqrt{d})$, where $d$ is the gradient dimension. This error term 
induces a high penalty for high dimensional models and thus cannot be applied to DNNs. Recently, \citep{diakonikolas2016robust} proposed a robust mean estimator with \emph{dimension-free} error 
for general types of corruptions. \citep{diakonikolas2019sever} achieves an error rate of $\mathcal{O}(\sqrt{\epsilon})$ for general corruption. This begs the question whether it is possible to further improve the $\mathcal{O}(\sqrt{\epsilon})$ error rate 
if we consider only corrupted supervision. 

To motivate our main algorithm (Alg.~\ref{alg:ro4label}), we first introduce and investigate Alg.~\ref{alg:ro_general} for general corruption with dimension-dependent error. The algorithm excludes data points with large gradient norms and uses the empirical mean of the remaining points to update the gradient. Cor.~\ref{coro:ro} below describes its robustness property.


\setlength{\textfloatsep}{-1pt}
	

  \begin{algorithm}
	\small
	\SetAlgoLined
	\textbf{input: } dataset $\rmD_x, \rmD_y^{\epsilon}$ with corrupted supervision, learning rate $\gamma_t$;\\
	\Return{{model parameter $\theta$};}\\
    \For{$t = 1$ to maxiter}{
        Randomly sample a mini-batch $\rmM$ from $\rmD_x, \rmD_y^{\epsilon}$\\
       Calculate the individual gradient $\tilde{\rmG}$ for $\rmM$\\
       For each row $\rvz_i$ in $\rmG$, calculate the l2 norm $\norm{\rvz_i}$ \\
       Choose the $\epsilon$-fraction rows with large $\norm{\rvz_i}$\\
	    Remove those selected rows, and return the empirical mean of the rest points as $\hat{\mu}$.\\
	    Update model $\mathbf{\theta}_{t+1} = \mathbf{\theta}_{t} - \gamma_{t}\hat{\mu}$
    }
	\caption{(\highlight{PRL(G)}) Provable Robust Learning for General Corrupted Data }
	\label{alg:ro_general}
\end{algorithm}

\begin{corollary}[Robust Optimization For Corrupted Data]
	\label{coro:ro}
	 Given the assumptions in Theorem~\ref{theo:convergence}, applying Algorithm~\ref{alg:ro_general} to  $\epsilon$-fraction corrupted data yields $\min_{t \in [T]}\expectation{\norm{\nabla \phi(\rvx_t)}} = \mathcal{O}(\epsilon \sqrt{d})$ for large enough $T$, where $d$ is the number of the parameters.
\end{corollary}

\begin{remark}
The term $\sqrt{d}$ is due to the upper bound of $d$-dimensional gradient norm of clean data. 
The term can be removed if we assume the gradient norm is uniformly bounded by $L$. However, this assumption is too strong for robust gradient estimation. We will show that later that the assumption can be relaxed (i.e. bounded maximum singular value of gradient) under the corrupted supervision setting.
\end{remark}

The error bound in the above corollary has several practical issues. First, the bound 
grows with increasing dimensionality, and thus, is prohibitive when working with DNNs, which have extremely large gradient dimensions due to their massive number of parameters. 
Even though one can improve the factor $\sqrt{\epsilon}$~\cite{diakonikolas2019sever} to $\epsilon$, the results remain impractical compared to the dimension-free $\mathcal{O}(\sqrt{\epsilon})$ guarantee in \citep{diakonikolas2019sever}, since above bound involves the dimension related term $\sqrt{d}$. 

Efficiency is another main limitation of Alg.~\ref{alg:ro_general}  since it requires computing individual gradients. Although there are advanced methods available to obtain the individual gradient, e.g., \citep{goodfellow2015efficient}, they are still relatively slow compared to the commonly used back-propagation algorithm. Moreover, many of them are not compatible with other  components of DNN such as batch normalization (BN). Since the individual gradients are not independent within the BN, they 
will lose the benefits of parallelization.
We will show below that the above issues can be addressed under the corrupted supervision setting and propose a practical solution that easily scales for DNNs.

\subsection{Robust Gradient Estimation for One Dimensional Corrupted Supervision}
In this section, we show that the robustness bound in  Cor.~\ref{coro:ro} can be improved if we assume the corruption comes from the supervision only. In addition, by fully exploiting the gradient structure of the corrupted supervision, our algorithm is much more efficient and is compatible with batch normalization. We begin with a  1-dimensional supervision setting (e.g., binary classification or single-target regression) to illustrate this intuition and will extend it more general settings in the next section.
Consider a supervised learning problem with input features $\rmX \in \mathbb{R}^{n \times p}$ and supervision $\mathbf{y} \in \mathbb{R}^{n}$. 
The goal is to learn a function $f$, parameterized by $\theta \in \mathbb{R}^d$, by minimizing the following loss $\min_{\theta}\sum_{i=1}^{n}\phi_i = \min_{\theta}\sum_{i=1}^{n}\mathcal{L}(y_i, f(\mathbf{x}_i, \theta))$. 
The gradient for a data point $i$ is $\nabla_{\theta}\phi_i = \tfrac{\partial l_i}{\partial f_i}\tfrac{\partial f_i}{\partial \theta} = \alpha_i \rvg_i$. 

In general, if the corrupted gradients drive the gradient estimation away from the clean gradient, they are either large in magnitude or systematically change the direction of the gradient~\cite{diakonikolas2019sever}.
However, our key observation is that, when only the supervision is corrupted, the corruption contributes only to the term $\alpha_i = \tfrac{\partial l_i}{\partial f_i}$, which is a scalar in the one-dimensional setting. 
In other words, given the clean gradient of $i^{th}$ point, $g_i \in \mathbb{R}^{d}$, the corrupted supervision only re-scales the gradient vector, changing the gradient from $\alpha_i \rvg_i$ to $\delta_i \rvg_i$, where $\delta_i = \tfrac{\partial l_i^{\epsilon}}{\partial f_i}$. As such, it is unlikely for the corrupted supervision to systematically change the gradient direction.


The fact that corrupted supervision re-scales the clean gradient can be exploited to reshape the robust optimization problem. Suppose we update our model in each iteration by $\theta^{+} = \theta - \gamma \mu(\rmG)$, where $\mu(\cdot)$ denotes the empirical mean function and $\mathbf{G} = [\nabla_{\theta}\phi_1^{T},\dots,\nabla_{\theta}\phi_m^{T}] \in \mathbb{R}^{m \times d}$ is the gradient matrix for a  mini-batch of size $m$. We consider the following problem:

\begin{problem}[Robust Gradient Estimation for One Dimensional Corrupted Supervision]
	\label{prob:rme}
Given a clean gradient matrix $\mathbf{G} \in \mathbb{R}^{m \times d}$, 
an $\epsilon$-corrupted matrix $\mathbf{\tilde{G}}$ with at most $\epsilon$-fraction rows are corrupted from $\alpha_i \rvg_i$ to $\delta_i \rvg_i$, design an algorithm $\mathcal{A}: \mathbb{R}^{m \times d} \rightarrow \mathbb{R}^d$ that minimizes $\norm{\mu(\mathbf{G})-\mathcal{A}(\mathbf{\tilde{G}})}$.
\end{problem}
Note that when $\norm{\delta_i}$ is large, the corrupted gradient will have a large effect on the empirical mean, and if $\norm{\delta_i}$ is small, the corrupted gradient will have a limited effect on the empirical mean. 
This motivates us to develop an algorithm that filters out data points by the loss layer gradient $\norm{\tfrac{\partial l_i}{\partial f_i}}$. 
If the norm of the loss layer gradient of a data point is large (in one-dimensional  case, this gradient reduces to a scalar and the norm becomes its absolute value), we exclude the data point when computing the empirical mean of gradients for this iteration. %
Note that this algorithm is applicable to both regression and classification problems.
Especially, when using the mean squared error (MSE) loss for regression, its gradient norm is exactly the loss itself, and the algorithm reduces to self-paced learning~\cite{kumar2010self} or trim loss~\cite{shen2019learning}. 
We summarize the procedure in Algorithm~\ref{alg:ro4label} and will extend it to the more general multi-dimensional case in the next section. 

  \begin{algorithm}
	\small
	\SetAlgoLined
	\textbf{input: } dataset $\rmD_x, \rmD_y^{\epsilon}$ with corrupted supervision, learning rate $\gamma_t$;\\
	\Return{{model parameter $\theta$};}\;
    \For{$t = 1$ to maxiter}{
        Randomly sample a mini-batch $\rmM$ from $\rmD_x, \rmD_y^{\epsilon}$\\
        Compute the predicted label $\hat{\rmY}$ from $\rmM$\\
        Calculate the gradient norm for the loss layer, (e.g., $\norm{\hat{\rvy} - \rvy}$ for mean square error or cross entropy)\\
        $\tilde{\rmM} \leftarrow \rmM - \rmM_{\tau}$, where $\rmM_{\tau}$ is the top-$\tau$ fraction of data points with largest $\norm{\hat{\rvy} - \rvy}$\\
        Update model $\theta_{t+1} = \theta_{t} - \gamma_{t}\hat{\mu}$, where $\hat{\mu}$ is the empirical mean of $\tilde{\rmM}$
    }
	\caption{(\highlight{PRL(L)}) Efficient Provable Robust Learning for Corrupted Supervision}
	\label{alg:ro4label}
\end{algorithm}
\subsection{Extension to Multi-Dimensional Corrupted Supervision}

To extend our approach to multi-dimensional case, let $q$  be the output dimension of $y$. The gradient for each data point $i$ is $\nabla_{\theta}\phi_i = \tfrac{\partial l_i}{\partial f_i}\tfrac{\partial f_i}{\partial \theta}$, where $\tfrac{\partial l_i}{\partial f_i} \in \mathbb{R}^{q}$ is the gradient of the loss with respect to model output, and $\tfrac{\partial f_i}{\partial \theta} \in \mathbb{R}^{q \times d}$ is the gradient of the model output with respect to model parameters. When the supervision is corrupted, the corruption affects the term $\tfrac{\partial l_i}{\partial f_i}$, which is now a vector. Let $\mathbf{\delta}_i = \tfrac{\partial l_i^\epsilon}{\partial f_i}\in \mathbb{R}^{q}$, $\mathbf{\alpha}_i = \tfrac{\partial l_i}{\partial f_i}\in \mathbb{R}^{q}$, $\mathbf{W}_i = \tfrac{\partial f_i}{\partial \theta} \in \mathbb{R}^{q \times d}$, and $m$ be the mini-batch size. Denote the clean gradient matrix as $\rmG \in \mathbb{R}^{m \times d}$, where the $i_{th}$ row of gradient matrix $\rvg_i = \alpha_i \rmW_i$. The multi-dimensional robust gradient estimation problem is defined as follows. 

\begin{problem}[Robust Gradient Estimation for Multi-Dimensional Corrupted Supervision]
	\label{prob:ml}
	Given a clean gradient matrix $\mathbf{G}$, 
	an $\epsilon$-corrupted matrix $\mathbf{\tilde{G}}$ with
	at most $\epsilon$-fraction rows corrupted from $\alpha_i \rmW_i$ to $\delta_i \rmW_i$, design an algorithm $\mathcal{A}: \mathbb{R}^{m \times d} \rightarrow \mathbb{R}^d$ that minimizes $\norm{\mu(\mathbf{G})-\mathcal{A}(\mathbf{\tilde{G}})}$.
\end{problem}

We begin our analysis by examining the effects of randomized filtering-base algorithms, i.e., using the empirical mean gradient of the random selected $(1-\epsilon)$-fraction subset to estimate clean averaged gradient.
Randomized filtering-based algorithm does not serve a practical robust learning approach, but its analysis leads to important insights into designing one.
We have the following lemma for \highlight{any randomized filtering-based algorithm} (proof is given in Appendix): 
\begin{lemma}[Gradient Estimation Error for Random Dropping $\epsilon$-fraction Data]
    Let $\tilde{\rmG} \in \mathbb{R}^{m \times d}$ be a corrupted matrix generated as in Problem~\ref{prob:ml}, and 
    $\rmG \in \mathbb{R}^{m \times d}$ be the original, clean gradient matrix. Suppose an arbitrary $(1-\epsilon)$-fraction rows are selected from $\tilde{\rmG}$ to form the matrix $\rmN \in \mathbb{R}^{n \times d}$. Let $\mu$ be the empirical mean function. Assume the clean gradient before loss layer has a bounded operator norm, i.e., $\norm{\rmW}_{op} \leq C$, the maximum clean gradient in loss layer $\max_{i \in \rmG}\norm{\alpha_i} = k$, and the maximum corrupted gradient in loss layer $\max_{i \in \rmN}\norm{\delta_i} = v$, then we have: 
	\begin{align*}
	\norm{\mu(\rmG) - \mu(\rmN)} \leq Ck\dfrac  {3\epsilon - 4\epsilon^2}{1-\epsilon} + Cv\dfrac{\epsilon}{1-\epsilon}.
	\end{align*}
	\label{lemma:randomdroperr}
\end{lemma}

Lemma \ref{lemma:randomdroperr} explains the factors affecting the robustness of filtering-based algorithm. Note that $v$ is the only term that is related to the corrupted supervision. If $v$ is large, then the bound is not safe since the right-hand side can be arbitrarily large (i.e. an adversary can change the supervision in such a way that $v$ becomes extremely large). Thus controlling the magnitude of $v$ provides a way to effectively reduce the bound. For example, if we manage to control $v \leq k$, then the bound is safe. This can be achieved by sorting the gradient norms at the loss layer, and then discarding the largest $\epsilon$-fraction data points. Motivated by Lemma \ref{lemma:randomdroperr}, we proposed Alg.~\ref{alg:ro4label}, whose robustness guarantee is given in Thm.~\ref{theo:mlrme} and Cor.~\ref{coro:ro4label}.
\begin{theorem}[Robust Gradient Estimation For Supervision Corruption]
	\label{theo:mlrme}
	Let $\tilde{\rmG}$ be a corrupted matrix generated as in Problem~\ref{prob:ml}, $q$  be the output dimension, and $\mu$ be the empirical mean of the clean gradient matrix $\rmG$. Assuming the maximum clean gradient before loss layer has bounded operator norm: $\norm{\rmW}_{op} \leq C$, then the output of gradient estimation in Algorithm~\ref{alg:ro4label}, $\hat{\mu}$, satisfies $\norm{\mu - \hat{\mu}} = \mathcal{O}(\epsilon\sqrt{q}) \approx \mathcal{O}(\epsilon)$.
\end{theorem}

Thm.~\ref{theo:mlrme} can be obtained from Lemma~\ref{lemma:randomdroperr} by substituting $v$ by $k$.
The following robustness guarantee can then be obtained by applying Thm.~\ref{theo:convergence}.

\begin{corollary}[Robust Optimization For Corrupted Supervision Data]
	\label{coro:ro4label}
	 Given the assumptions used in Thm.~\ref{theo:convergence}, applying Algorithm.~\ref{alg:ro4label} to any $\epsilon$-fraction supervision corrupted data, yields $\min_{t \in [T]}\expectation{\norm{\nabla \phi(\rvx_t)}} = \mathcal{O}(\epsilon \sqrt{q})$ for large enough $T$, where $q$ is the dimension of the supervision.
\end{corollary}

Comparing Cor.~\ref{coro:ro} and Cor.~\ref{coro:ro4label}, we see that when the corruption only comes from supervision, the dependence on $d$ is reduced to $q$, where $q \ll d$ in most deep learning problems. 

\subsection{Comparison against 
Other Robust Optimization Methods}
SEVER \citep{diakonikolas2019sever} provides state-of-the-art theoretical results for \emph{general corruptions}, with a promising $\mathcal{O}(\sqrt{\epsilon})$ dimension-free guarantee. Compared to \citet{diakonikolas2019sever}, we have two contributions:
\textbf{a)} When corruption comes only from the supervision, we show a better error rate if supervision dimension can be treated as a small constant.
\textbf{b)} Our algorithm can scale to DNNs while \citet{diakonikolas2019sever} cannot. This is especially critical as the DNN based models are currently state-of-the-art methods for noisy label learning problems.

Despite the impressive theoretical results in \citet{diakonikolas2019sever}, it cannot be applied to DNNs even with the current best hardware configuration. \citet{diakonikolas2019sever} used dimension-free robust mean estimation techniques to design the learning algorithm, while most robust mean estimation approaches rely on filtering data by computing the score of projection to the maximum singular vector. For example, the approach in \citet{diakonikolas2019sever} requires applying  expensive SVD on $n \times d$ individual gradient matrix, where $n$ is the sample size and $d$ is the number of parameters. This method works well for smaller datasets and smaller models when both $n$ and $d$ are small enough for current memory limitation. However, for DNNs, this matrix size is far beyond current GPU memory capability. For example, in our experiment, $n$ is 60,000 and $d$ is in the order of millions (network parameters). It is impractical to store 60,000 copies of networks in a single GPU card. In contrast, our algorithm does not need to store the full gradient matrix. By only considering the loss-layer gradient norm, it can be easily extended to DNNs, and we show that this simple strategy works well in both theory and challenging empirical tasks.

We note that better robustness guarantee can be achieved in linear \citep{bhatia2015robust, bhatia2017consistent} or convex \citep{prasad2018robust} cases, but they cannot be directly applied to DNNs. 

The strongest assumption behind our proof is that the maximum singular value of the gradient before loss layer is bounded, which is similar to the one used in \citet{diakonikolas2019sever}. We also treat the clean gradient loss layer norm ($k$ in Lemma~\ref{lemma:randomdroperr}) as a constant, which is particularly true for DNNs due to their overparameterization. In practice, our algorithm slowly increase the dropping ratio $\tau$ at first few epochs, which guarantees that $k$ is a small number.

\subsection{Relationship to Self-Paced Learning (SPL)}
Many state-of-the-art methods with noisy labels depend on the SPL \citep{han2018co, song2019selfie, yu2019does, shen2019learning, wei2020combating, sun2020robust}. At first glance, our method looks very similar to SPL. Instead of keeping data points with small gradient norms, SPL tries to keep those points with small loss. The gradient norm and loss function are related via the famous Polyak-Łojasiewicz (PL) condition. The PL condition assumes there exists some constant $s > 0$ such that $\forall \rvx: \frac{1}{2}\|\nabla \phi(\rvx)\|^{2} \geq s\left(\phi(\rvx)-\phi^{*}\right)$.
As we can see, when the neural network is highly over-parameterized, $\phi^*$ can be assumed to be equal across different samples since the neural networks can achieve zero training loss \citep{zhang2016understanding}. By sorting the error $\phi(\rvx_i)$ for every data point, SPL is actually sorting the lower bound of the gradient norm if the PL condition holds. However, the ranking of gradient norm and the ranking of the loss can be very different since there is no guarantee that the gradient norm is monotonically increasing with the loss value. 

\begin{figure}[]
	\centering
	\begin{subfigure}{0.45\textwidth}
		\includegraphics[width=\linewidth]{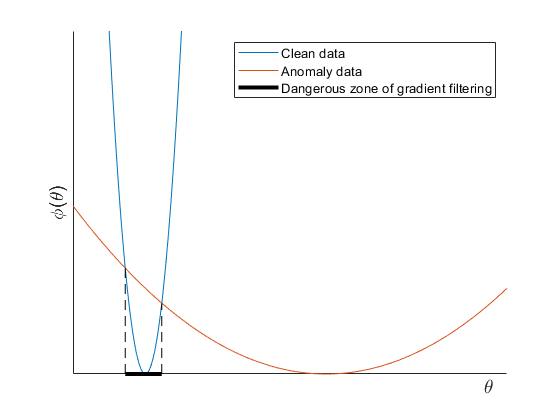}
		\subcaption{When gradient filtering method failed to pick out right corrupted data, the remaining corrupted data is relatively smooth, thus has limited impact on overall loss surface.}
		\label{fig:dangerzonegradient}
	\end{subfigure}
	\begin{subfigure}{0.45\textwidth}
		\includegraphics[width=\linewidth]{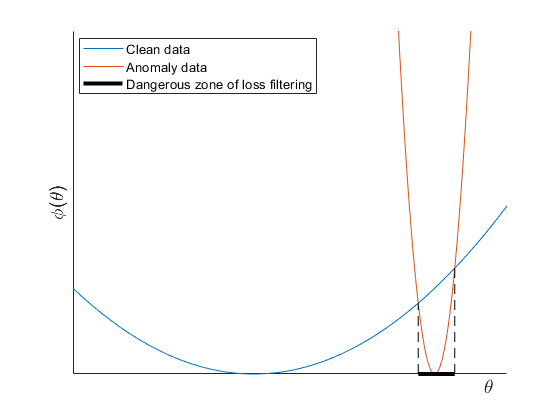}
		\subcaption{When loss filtering method failed to pick out right corrupted data, the remaining corrupted data could be extremely sharp, thus has large impact on overall loss surface.}
		\label{fig:dangerzoneloss}
	\end{subfigure}

	\caption{Geometric illustration of the difference between loss filtering and gradient norm filtering.
		\label{fig:twocurve}
	}
\end{figure}

Here we show that the monotonic relationship can be easily violated even for the simple square loss function.  One easy counter-example is $\phi(x_1,x_2) = 0.5 x_1^2 + 50 x_2^2$. Take two points (1000, 1) and (495, -49.5), we will find the monotonic relationship does not hold for these two points. \citet{nocedal2002behavior} showed that the monotonic relationship holds for \highlight{square loss} (i.e. $\phi(\rvx) = \tfrac{1}{2}(\rvx-\rvx^*)^{T}\rmQ(\rvx-\rvx^*)$ ) if the condition number of Q is smaller than $3 + 2\sqrt{2}$, which is quite a strong assumption especially when $\rvx$ is in high-dimension. If we consider the more general type of loss function (e.g., neural network), the assumptions on condition number should only be stronger, thus breaking the monotonic relationship. Thus, although SPL sorts the lower bound of the gradient norm under mild assumptions, our algorithm is significantly different from SPL and its variations.

We also provide an illustration as to why SPL is not robust from the loss landscape perspective in figure \ref{fig:twocurve}. In order to have a more intuitive understanding of our algorithm, we could look at the Figure \ref{fig:dangerzonegradient} and \ref{fig:dangerzoneloss}. Since we are in the agnostic label corruption setting, it is difficult to filtering out the correct corrupted data. We showed two situations when loss filtering failed and gradient filtering failed.  As we could see that when loss filtering method failed, the remaining corrupted data could have large impact on the overall loss surface while when gradient filtering method failed, the remaining corrupted data only have limited impact on the overall loss surface, thus gaining robustness.

Next, we discuss the relationship between SPL and Algorithm \ref{alg:ro4label} under corrupted supervision. SPL has the same form as Algorithm \ref{alg:ro4label} when we are using mean square error to perform regression tasks since the loss layer gradient norm is equal to loss itself. However, in classification, Algorithm \ref{alg:ro4label} is different from the SPL. In order to better understand the algorithm, we further analyze the difference between SPL and our algorithm for cross-entropy loss.

For cross entropy, denote the output logit as $\rvo$, we have
\begin{align*}
    H(\rvy_i, \rvf_i) =-\langle\rvy_i, \log(\softmax(\rvo_i)) \rangle = -\langle\rvy_i, \log(\rvf_i) \rangle.
\end{align*}
 The gradient norm of cross entropy with respect to $\rvo_i$ is:
$\frac{\partial H_i}{\partial \rvo_i} = \rvy_i - \softmax(\rvo_i) = \rvf_i - \rvy_i$. Thus, the gradient norm of loss layer is the MSE between $\rvy_i$ and $\rvf_i$. Next, we investigate when MSE and cross entropy give non-monotonic relationship. For simplicity, we only consider the sufficient condition for the non-monotonic relationship, which is given by Lemma \ref{lem:mse}.          


\begin{lemma}
	Let $\mathbf{y} \in \mathbb{R}^q$, where $\rvy_k=1$ and $\rvy_i=0$ for $i \neq k$. Suppose $\alpha$ and $\beta$ are two $q$-dimensional vectors in probability simplex. Without loss of generality, suppose $\alpha$ has a  smaller cross entropy loss and $\alpha_k \geq \beta_k$, then the sufficient condition for $\norm{\mathbf{\alpha}-\mathbf{y}} \geq \norm{\beta - \rvy}$ is $\Var_{i \neq k}(\{\alpha_i\}) - \Var_{i \neq k}(\{\beta_i\})  \geq \frac{q}{(q-1)^2}\left((\alpha_k-\beta_k)(2-\alpha_k-\beta_k)\right)$
	\label{lem:mse}
\end{lemma}

As $\alpha_k \geq \beta_k$, the term on right-hand-side of the inequality is non-negative. Thus, when MSE generates a result that differs from cross-entropy, the variance in the probability vector of the non-true class for the discarded data point is larger. For example, consider the ground-truth vector $\rvy=[0,1,0,0,0, 0, 0,0 ,0, 0]$, and two prediction vectors, $\alpha=[0.08, 0.28, 0.08, 0.08, 0.08, 0.08, 0.08, 0.08, 0.08, 0.08]$ and $\beta=[0.1, 0.3, 0.34, 0.05, 0.05, 0.1, 0.03, 0.03, 0, 0]$. $\alpha$ has a smaller MSE loss while $\beta$ has a smaller cross-entropy loss. $\beta$ will more likely be noisy data since it has two relatively large values of 0.3 and 0.34. Since cross entropy loss considers only one dimension, corresponding to the ground truth label, it cannot detect such a situation. Compared to cross-entropy, the gradient (mse loss) considers all dimensions, and thus, will consider the distribution of the overall prediction.



\subsection{Combining with Co-teaching Style Training}
Co-teaching \citep{han2018co} is one of the state-of-the-art deep methods for learning with noisy labels. Motivated by Co-teaching, we propose  \highlight{Co-PRL(L)}, which has the same framework as co-teaching but uses the loss-layer gradient to select the data. The key difference between \highlight{Co-PRL(L)} and algorithm \ref{alg:ro4label} is that in \highlight{Co-PRL(L)}, we optimize two network by \highlight{PRL(L)}. Also in every iteration, two networks will exchange the selected data to update their own parameters. The algorithm is in \ref{alg:Co-GIFilter}.
  \begin{algorithm}[]
	\small
	\SetAlgoLined
	\textbf{input: } initialize $w_f$ and $w_g$, learning rate $\eta$, fixed $\tau$, epoch $T_k$ and $T_{max}$, iterations $N_{max}$\\
	\Return{model parameter $w_f$ and $w_g$}\;
	\For{$T = 1, 2, ..., T_{max}$}{
		\For{$N = 1,..., N_{max}$}{
			random sample a minibatch $\rmM$ from $\rmD_x, \rmD_y^{\epsilon}$  \hspace{1cm}(noisy dataset)\\
			get the predicted label $\hat{\rmY}_f$ and $\hat{\rmY}_g$ from $\rmM$ by $w_f$. $w_g$\\
			calculate the individual loss $l_f = \mathcal{L}(\rmY, \hat{\rmY}_f)$, $l_g = \mathcal{L}(\rmY, \hat{\rmY}_g)$\\
			calculate the gradient norm of loss layer $score_f = \norm{\dfrac{\partial l_f}{\partial \hat{\rvy}_f}}$, $score_g = \norm{\dfrac{\partial l_g}{\partial \hat{\rvy}_g}}$.\\ 
			sample $R(T)\%$ small-loss-layer-gradient-norm instances by $score_f$ and $score_g$ to get $\rmN_f$, $\rmN_g$\\
			update $w_f = w_f - \eta \nabla_{w_f}\mathcal{L}(\rmN_f, w_f)$, $w_g = w_g - \eta \nabla_{w_g}\mathcal{L}(\rmN_g, w_g)$ \hspace{1cm}(selected dataset)\\
			update model $\rvx_{t+1} = \rvx_{t} - \gamma_{t}\hat{\mu}$	
		}
		\textbf{Update} $R(T) = 1 - \min\left\{\dfrac{T}{T_k}\tau, \tau\right\}$
	}
	\caption{Co-PRL(L)}
	\label{alg:Co-GIFilter}
\end{algorithm}

\vspace{-0.3cm}
\section{Experimental Results}

\begin{table*}[t]
    \centering
	\resizebox{1\linewidth}{!}{
	\begin{tabular}{|l|l|l|l|l|l||l|l|l|}
		\hline
		Corruption   & Standard     & Normclip      & Huber       & Min-sgd     & Ignormclip   & PRL(G)             & PRL(L)          & Co-PRL(L)       \\ \hline
		linadv: 10   & -2.33±0.84   & -2.22±0.74    & 0.868±0.01 & 0.103±0.03  & 0.68±0.07   & \textbf{0.876±0.01} & \textbf{0.876±0.01}         & \textbf{0.876±0.01}        \\ \hline
		linadv: 20   & -8.65±2.1    & -8.55±2.2     & 0.817±0.015 & 0.120±0.02   & 0.367±0.28   & \textbf{0.871±0.01} & 0.869±0.01          & 0.869±0.01          \\ \hline
		linadv: 30   & -18.529±4.04 & -19.185±4.31  & 0.592±0.07  & 0.146±0.03 & -0.944±0.51  & \textbf{0.865±0.01}  & 0.861±0.01           & 0.860±0.01           \\ \hline
		linadv: 40   & -32.22±6.32  & -32.75±7.07   & -2.529±1.22 & 0.180±0.01 & -1.60 ± 0.80 & \textbf{0.857± 0.01} & 0.847±0.02           & 0.847±0.02           \\ \hline\hline
		signflip: 10 & 0.800±0.02    & 0.798±0.03   & 0.857±0.01 & 0.110±0.04 & 0.846±0.01  & 0.877±0.01          & 0.878±0.01          & \textbf{0.879±0.01} \\ \hline
		signflip: 20 & 0.641±0.05  & 0.638±0.04   & 0.786±0.02 & 0.105±0.07 & 0.82±0.02   & 0.875±0.01         & 0.875±0.01          & \textbf{0.877±0.01} \\ \hline
		signflip: 30 & 0.422±0.04   & 0.421±0.04    & 0.629±0.03  & 0.124±0.05  & 0.795±0.02   & 0.871±0.01           & 0.873±0.01           & \textbf{0.875±0.01}  \\ \hline
		signflip: 40 & 0.193±0.043  & 0.190±0.04    & 0.379±0.05  & -0.028±0.25  & 0.759±0.01   & \textbf{0.872±0.01} & \textbf{0.872±0.01}  & 0.871±0.01           \\ \hline \hline
		uninoise: 10 & 0.845±0.01  & 0.844±0.01   & 0.875±0.01 & 0.103±0.03 & 0.859±0.01  & 0.879±0.01          & \textbf{0.881±0.01} & \textbf{0.881±0.01} \\ \hline
		uninoise: 20 & 0.798±0.02   & 0.795±0.02    & 0.865±0.01  & 0.120±0.02  & 0.844±0.01   & 0.878±0.01           & \textbf{0.880±0.01}  & \textbf{0.880±0.01}  \\ \hline
		uninoise: 30 & 0.728±0.02   & 0.725±0.02    & 0.847±0.01  & 0.146±0.03  & 0.831±0.01   & 0.878±0.01           & \textbf{0.879±0.01}           & \textbf{0.879±0.01}  \\ \hline
		uninoise: 40 & 0.656±0.02   & 0.654±0.02    & 0.825±0.01  & 0.180±0.01  & 0.821±0.01   & 0.876± 0.01          & \textbf{0.878±0.01}          & \textbf{0.878±0.01}  \\ \hline\hline
		pairflip: 10 & 0.852±0.02  & 0.851±0.02    & 0.870±0.01  & 0.110±0.04 & 0.867±0.01  & 0.877±0.01          & 0.876±0.01          & \textbf{0.878±0.01} \\ \hline 
		pairflip: 20 & 0.784±0.03   & 0.783±0.03    & 0.841±0.02  & 0.120±0.03  & 0.849±0.01   & \textbf{0.874±0.01}  & 0.873±0.01           & \textbf{0.874±0.01}  \\ \hline
		pairflip: 30 & 0.688±0.04   & 0.686±0.04    & 0.770±0.02  & 0.133±0.02  & 0.828±0.01   & 0.870±0.01           & 0.872±0.01           & \textbf{0.873±0.01}  \\ \hline
		pairflip: 40 & 0.556±0.06   & 0.553±0.06    & 0.642±0.06  & 0.134±0.03  & 0.810±0.02   & 0.863±0.01           & \textbf{0.870±0.01} & \textbf{0.870±0.01}  \\ \hline \hline
		mixture: 10  & -0.212±0.6   & -0.010±0.48 & 0.873±0.01 & 0.101±0.03  & 0.861±0.01  & 0.878±0.01          & \textbf{0.880±0.01}  & \textbf{0.880±0.01}  \\ \hline
		mixture: 20  & -0.404±0.68  & -0.463±0.67   & 0.855±0.01  & 0.119±0.03  & 0.855±0.01   & 0.877±0.01           & 0.878±0.01           & \textbf{0.879±0.01}  \\ \hline
		mixture: 30  & -0.716±0.57  & -0.824±0.39   & 0.823±0.01  & 0.148±0.02 & 0.847±0.01   & 0.875±0.01           & 0.877±0.01           & \textbf{0.878±0.01}  \\ \hline
		mixture: 40  & -3.130±1.51   & -2.69±0.84    & 0.763±0.01  & 0.175±0.02  & 0.835±0.01   & 0.872±0.01           & 0.875 ±0.01          & \textbf{0.876±0.01}  \\ \hline
	\end{tabular}
	}
	\caption{R-square on CelebA clean testing data, and the standard deviation is from last ten epochs and 5 random seeds. }
	\label{tab:regression}
\end{table*}

\begin{table*}[t]
\centering
\resizebox{1\linewidth}{!}{
\begin{tabular}{|l|l|l|l|l|l||l|l||l|l|}
\hline
Corruption   & Standard     & Normclip     & Bootstrap    & Decouple     & Min-sgd      & SPL          & PRL(L)  & Co-teaching  & Co-PRL(L)        \\ \hline
CF10-sym-30  & 63.22±0.18 & 62.41±0.06 & 63.67±0.24 & 70.73±0.51 & 13.31±2.24 & 77.77±0.34 & 79.40±0.19 & 79.90±0.13 & \textbf{80.05±0.12} \\ \hline
CF10-sym-50  & 44.63±0.18 & 43.99±0.28 & 46.13±0.18 & 57.48±1.98 & 13.33±2.85 & 72.22±0.15 & 74.17±0.15 & 74.25±0.41 & \textbf{75.43±0.09} \\ \hline
CF10-sym-70  & 24.12±0.09 & 24.17±0.37 & 25.13±0.39 & 40.11±4.62 & 9.08±0.94  & 56.19±0.33 & 58.36±0.62 & 58.41±0.33 & \textbf{60.26±0.42} \\ \hline \hline
CF10-pf-25   & 68.34±0.30 & 67.92±0.43 & 68.71±0.32 & 75.59±0.35 & 10.45±0.60 & 75.79±0.44 & 80.54±0.07 & 80.18±0.21 & \textbf{81.51±0.13} \\ \hline
CF10-pf-35   & 58.68±0.28 & 58.27±0.18 & 58.19±0.12 & 66.38±0.44 & 12.29±1.92 & 70.40±0.27 & 77.61±0.35 & 77.97±0.03 & \textbf{79.01±0.14} \\ \hline
CF10-pf-45   & 48.05±0.25 & 48.03±0.54 & 47.84±0.32 & 51.54±0.81 & 10.94±1.28 & 58.95±0.59 & 71.42±0.24 & 72.43±0.31 & \textbf{73.78±0.17} \\ \hline \hline
CF100-sym-30 & 32.83±0.39 & 32.10±0.64 & 34.47±0.22 & 32.95±0.44 & 2.94±0.61  & 44.37±0.44 & 46.40±0.18 & 45.02±0.29 & \textbf{47.51±0.47} \\ \hline
CF100-sym-50 & 20.47±0.44 & 19.73±0.29 & 21.59±0.44 & 21.02±0.36 & 2.35±0.45  & 37.89±0.16 & 38.38±0.65 & 38.79±0.33 & \textbf{40.64±0.11} \\ \hline
CF100-sym-70 & 9.93±0.07  & 9.93±0.23  & 10.59±0.17 & 12.55±0.46 & 2.32±0.24  & 24.10±0.44 & 25.38±0.56 & 24.94±0.53 & \textbf{27.27±0.01} \\ \hline \hline
CF100-pf-25 & 40.37±0.55 & 39.34±0.35 & 40.22±0.37 & 39.43±0.27 & 2.62±0.26  & 40.48±0.72 & 47.57±0.37 & 42.97±0.10 & \textbf{48.06±0.26} \\ \hline
CF100-pf-35 & 34.07±0.19 & 32.88±0.10 & 34.53±0.23 & 33.14±0.07 & 2.30±0.07  & 34.17±0.46 & 43.32±0.16 & 36.69±0.23 & \textbf{44.08±0.33} \\ \hline
CF100-pf-45 & 27.66±0.50 & 27.35±0.61 & 27.56±0.23 & 26.83±0.41 & 2.55±0.52  & 27.55±0.66 & 33.31±0.10 & 29.71±0.20 & \textbf{34.43±0.05} \\ \hline
\end{tabular}
}
	\caption{Classification accuracy for clean testing data on CIFAR10 and CIFAR100 with training on \highlight{symmetric} and \highlight{pairflip} label corruption. The standard deviation is from last ten epochs and 3 random seeds. }
	\label{tab:classification}
\end{table*}

We have performed our experiments on various benchmark regression and classification datasets. We compare \highlight{PRL(G)}(Algorithm \ref{alg:ro_general}), \highlight{PRL(L)} (Algorithm \ref{alg:ro4label}), and \highlight{Co-PRL(L)} (Algorithm \ref{alg:Co-GIFilter}) to the following baselines. 
\begin{itemize}
    \item \highlight{Standard}: standard training without filtering data (mse for regression, cross entropy for classification);
    \item \highlight{Normclip}: standard training with norm clipping; \highlight{Huber}: standard training with huber loss (regression only);
    \item \highlight{Decouple}: decoupling network, update two networks by using their disagreement \citep{malach2017decoupling} (classification only);
    \item \highlight{Bootstrap}: It uses a weighted combination of predicted and original labels as the correct labels, and then perform back propagation \citep{reed2014training} (classification only);
    \item \highlight{Min-sgd}: choosing the smallest loss sample in minibatch to update model \citep{shah2020choosing};
    \item \highlight{SPL}~\cite{jiang2018mentornet}: self-paced learning (also known as the trimmed loss), dropping the data with large losses (same as \highlight{PRL(L)} in regression setting with MSE loss);
    \item \highlight{Ignormclip}: clipping individual gradient then average them to update model (regression only); 
    \item \highlight{Co-teaching}: collaboratively train a pair of SPL model and exchange selected data to another model~\citep{han2018co} (classification only).
\end{itemize}

Since it is hard to design experiments for \highlight{agnostic corrupted supervision}, 
we analyzed the performance on a broad  class 
of corrupted supervision settings: 
\begin{itemize}
    \item \highlight{linadv}: the corrupted supervision is generated by random wrong linear relationship of features: $\rmY_{\epsilon} = \rmX * \rmW_{\epsilon}$ (regression); 
    \item \highlight{signflip}: the supervision sign is flipped $\rmY_{\epsilon} = -\rmY$ (regression); 
    \item \highlight{uninoise}: random sampling from uniform distribution as corrupted supervision $\rmY_{\epsilon} \sim [-5, 5]$ (regression); 
    \item \highlight{mixture}: mixture of above types of corruptions (regression);
    \item \highlight{pairflip}: shuffle the coordinates (i.e. eyes to mouth in CelebA or cat to dog in CIFAR) (regression and classification); 
    \item \highlight{symmetric}: randomly assign wrong class label (classification).
\end{itemize}

For classification, we use accuracy as the evaluation metric, and R-square is used to evaluate regression experiments. We show the average evaluation score on testing data for the last 10 epochs. We also include the training curves to show how the testing evaluation metric changes during training phase. All experiments are repeated 5 times for regression experiments and 3 times for classification experiments. Main hyperparameters are showed in the  Table \ref{tab:hyper}. For Classification, we use the same hyperparameters in \cite{han2018co}. For CelebA, we use 3-layer fully connected network with 256 hidden nodes in hidden layer and leakly-relu as activation function.
\begin{table}[]
\centering
\begin{tabular}{|l|l|l|l|l|}
\hline
Data\textbackslash{}HyperParameter & BatchSize & Learning Rate & Optimizer & Momentum \\ \hline
CF-10                              & 128       & 0.001         & Adam      & 0.9      \\ \hline
CF-100                             & 128       & 0.001         & Adam      & 0.9      \\ \hline
CelebA                             & 512       & 0.0003        & Adam      & 0.9      \\ \hline
\end{tabular}
\caption{Main Hyperparmeters}
\label{tab:hyper}
\end{table}

\begin{figure}[t!]
	\centering
	\begin{subfigure}{0.24\textwidth}
		\includegraphics[width=\linewidth]{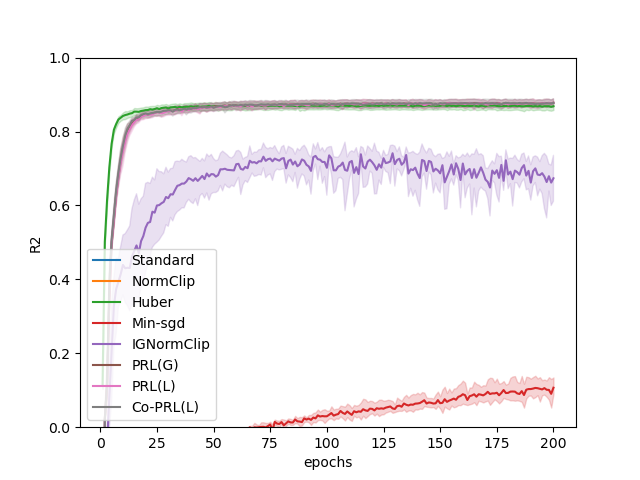}
		\subcaption{$\epsilon=0.1$ linadv noise}
	\end{subfigure}
	\begin{subfigure}{0.24\textwidth}
		\includegraphics[width=\linewidth]{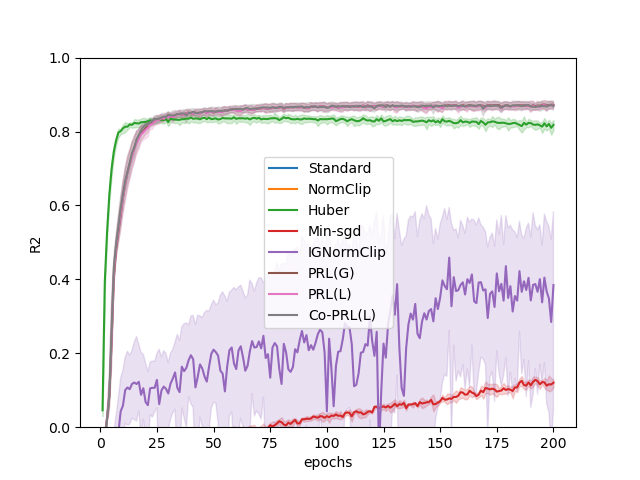}
		\subcaption{$\epsilon=0.2$ linadv noise}
	\end{subfigure}
	\begin{subfigure}{0.24\textwidth}
	\includegraphics[width=\linewidth]{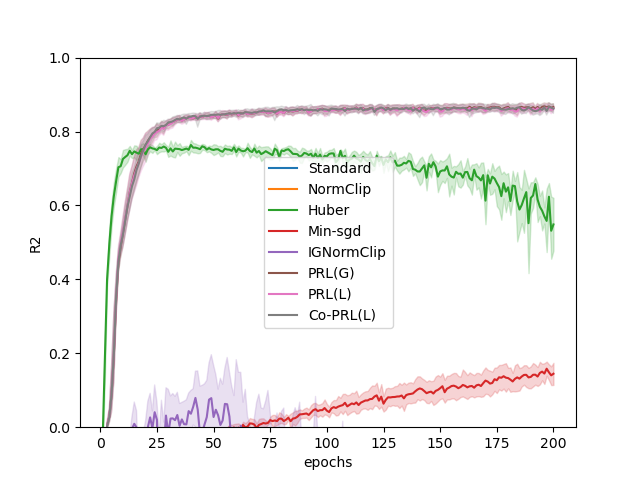}
	\subcaption{$\epsilon=0.3$ linadv noise}
\end{subfigure}
	\begin{subfigure}{0.24\textwidth}
	\includegraphics[width=\linewidth]{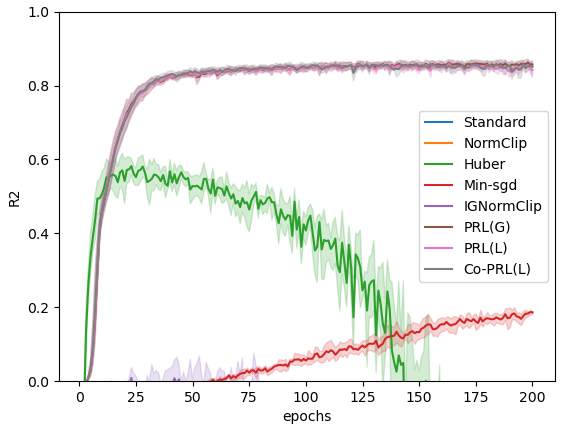}
	\subcaption{$\epsilon=0.4$ linadv noise}
\end{subfigure}
\qquad
	\begin{subfigure}{0.24\textwidth}
	\includegraphics[width=\linewidth]{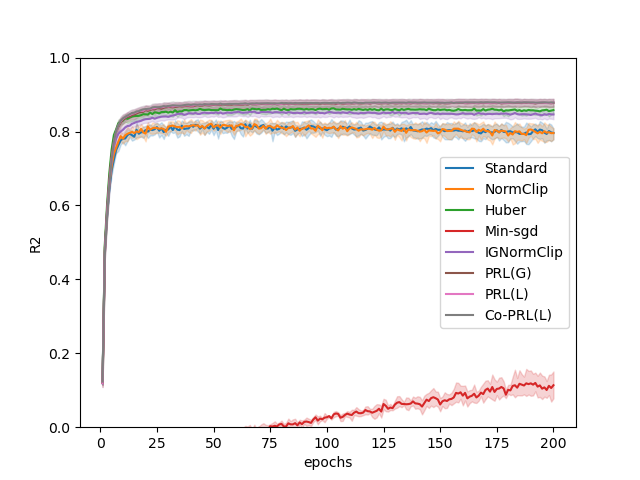}
	\subcaption{$\epsilon=0.1$ signflip noise}
\end{subfigure}
\begin{subfigure}{0.24\textwidth}
	\includegraphics[width=\linewidth]{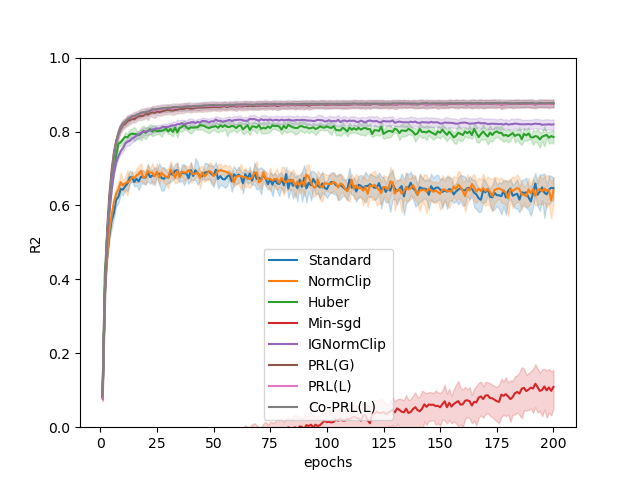}
	\subcaption{$\epsilon=0.2$ signflip noise}
\end{subfigure}
\begin{subfigure}{0.24\textwidth}
	\includegraphics[width=\linewidth]{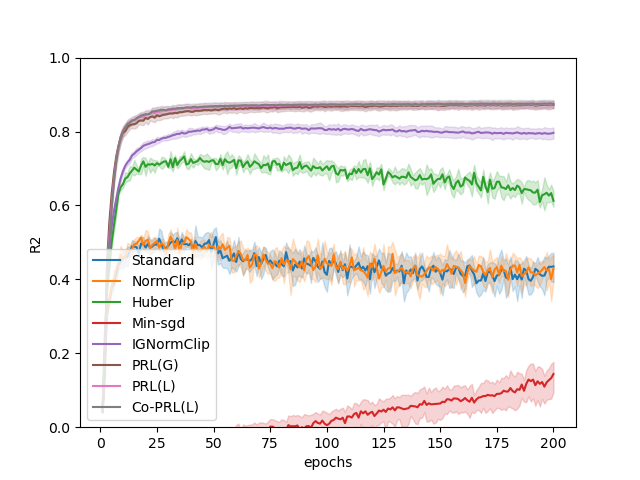}
	\subcaption{$\epsilon=0.3$ signflip noise}
\end{subfigure}
\begin{subfigure}{0.24\textwidth}
	\includegraphics[width=\linewidth]{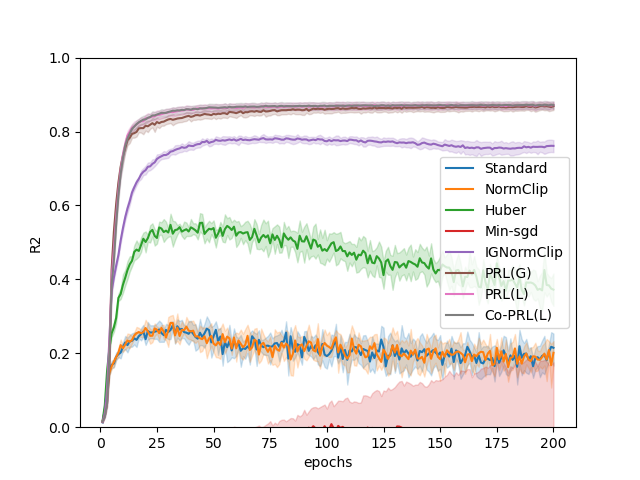}
	\subcaption{$\epsilon=0.4$ signflip noise}
\end{subfigure}
\qquad
	\begin{subfigure}{0.24\textwidth}
	\includegraphics[width=\linewidth]{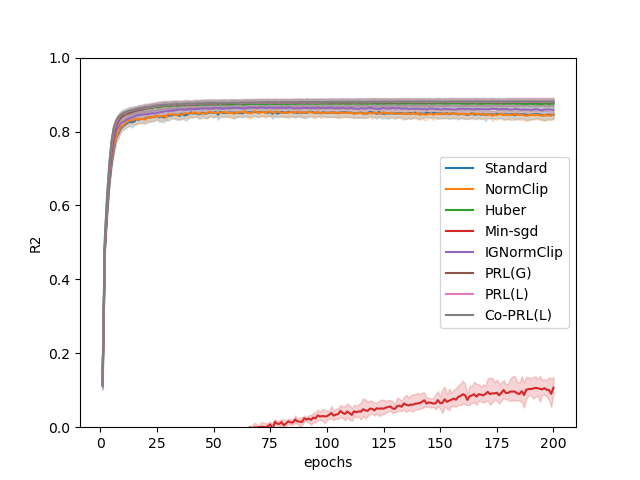}
	\subcaption{$\epsilon=0.1$ uninoise noise}
\end{subfigure}
\begin{subfigure}{0.24\textwidth}
	\includegraphics[width=\linewidth]{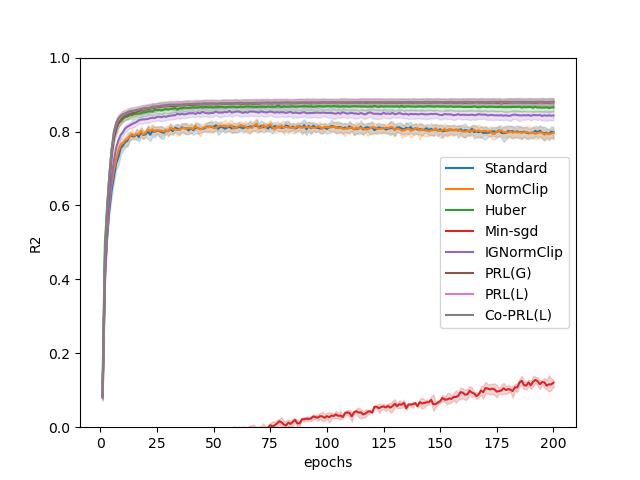}
	\subcaption{$\epsilon=0.2$ uninoise noise}
\end{subfigure}
\begin{subfigure}{0.24\textwidth}
	\includegraphics[width=\linewidth]{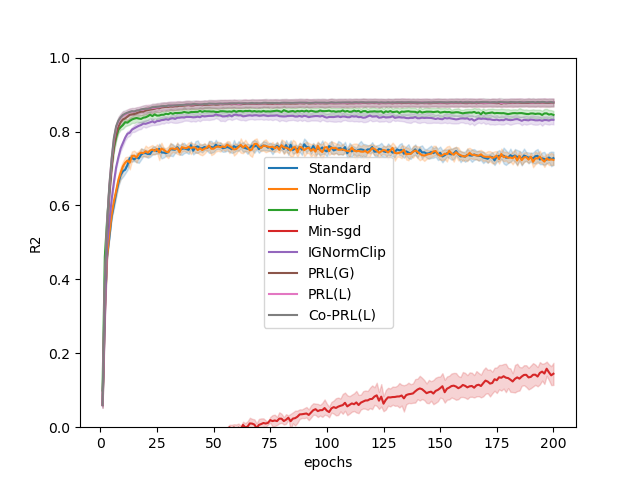}
	\subcaption{$\epsilon=0.3$ uninoise noise}
\end{subfigure}
\begin{subfigure}{0.24\textwidth}
	\includegraphics[width=\linewidth]{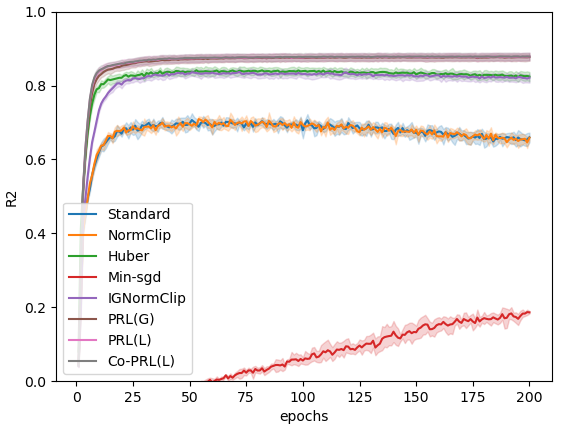}
	\subcaption{$\epsilon=0.4$ uninoise noise}
\end{subfigure}
\qquad
	\begin{subfigure}{0.24\textwidth}
	\includegraphics[width=\linewidth]{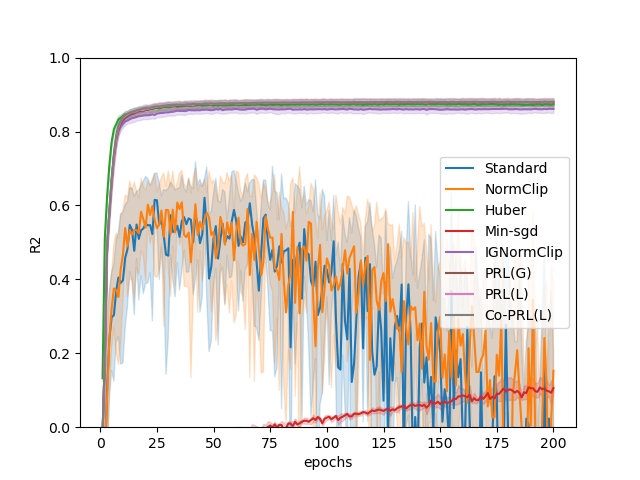}
	\subcaption{$\epsilon=0.1$ mixture noise}
\end{subfigure}
\begin{subfigure}{0.24\textwidth}
	\includegraphics[width=\linewidth]{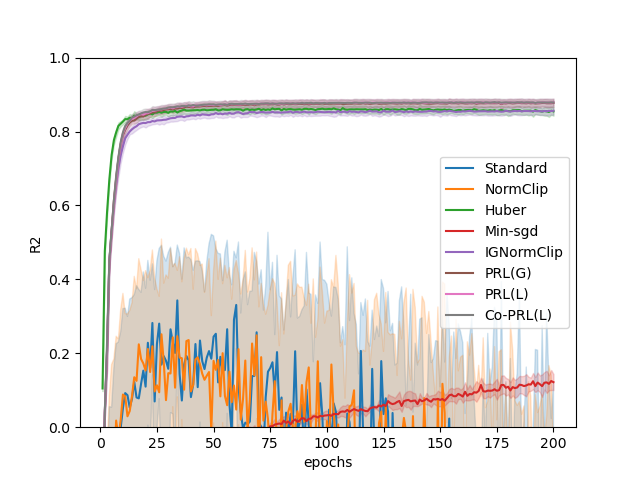}
	\subcaption{$\epsilon=0.2$ mixture noise}
\end{subfigure}
\begin{subfigure}{0.24\textwidth}
	\includegraphics[width=\linewidth]{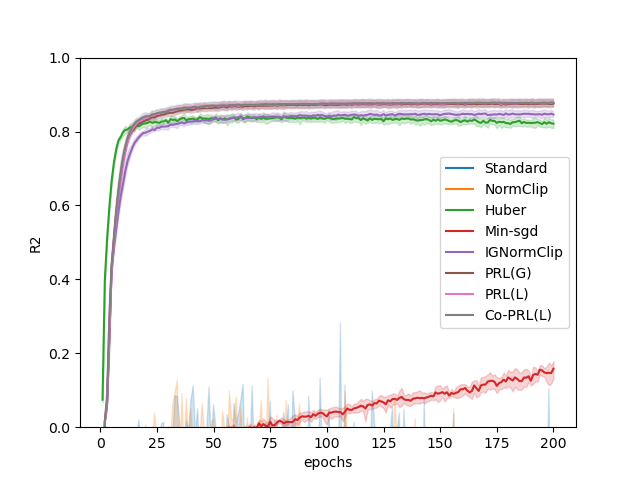}
	\subcaption{$\epsilon=0.3$ mixture noise}
\end{subfigure}
\begin{subfigure}{0.24\textwidth}
	\includegraphics[width=\linewidth]{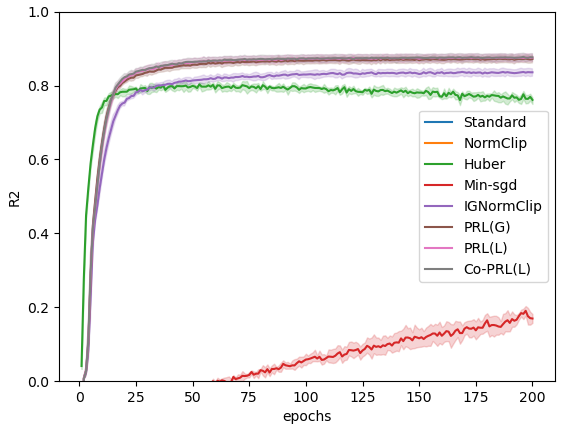}
	\subcaption{$\epsilon=0.4$ mixture noise}
\end{subfigure}
\qquad
	\begin{subfigure}{0.24\textwidth}
	\includegraphics[width=\linewidth]{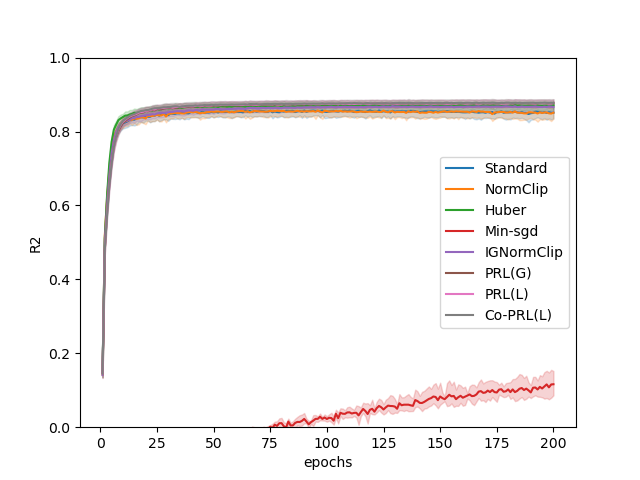}
	\subcaption{$\epsilon=0.1$ pairflip noise}
\end{subfigure}
\begin{subfigure}{0.24\textwidth}
	\includegraphics[width=\linewidth]{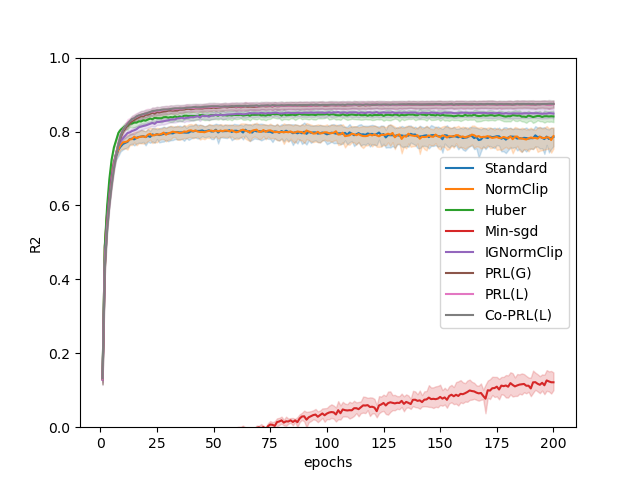}
	\subcaption{$\epsilon=0.2$ pairflip noise}
\end{subfigure}
\begin{subfigure}{0.24\textwidth}
	\includegraphics[width=\linewidth]{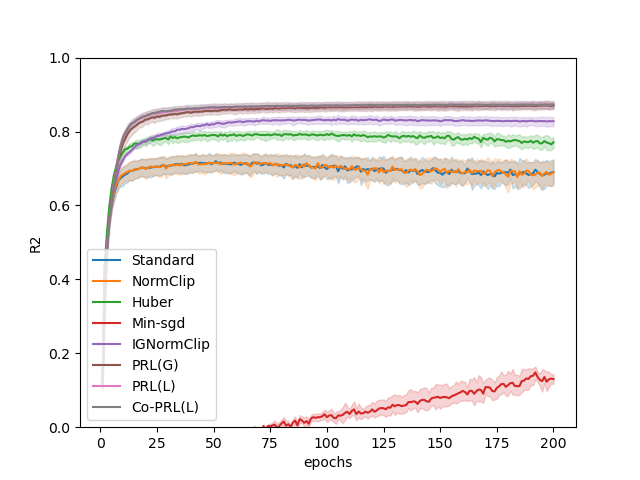}
	\subcaption{$\epsilon=0.3$ pairflip noise}
\end{subfigure}
\begin{subfigure}{0.24\textwidth}
	\includegraphics[width=\linewidth]{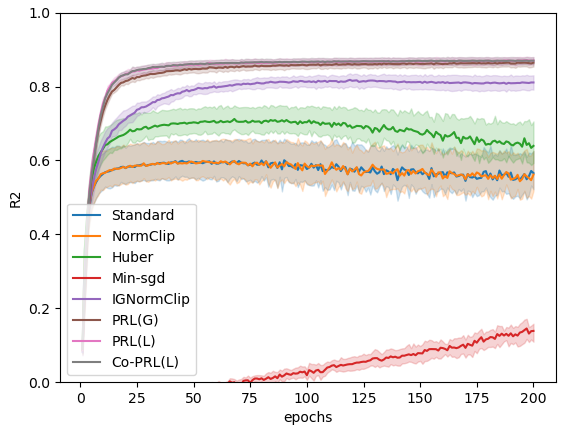}
	\subcaption{$\epsilon=0.4$ pairflip noise}
\end{subfigure}

\caption{CelebA Testing Curve During Training. X axis represents the epoch number, Y axis represents the testing r-square. In some experiment, there is no curve for Standard and NormClip since they gave negative r-square, which will effect the plotting scale. The shadow represents the confidence interval, which is calculated across 5 random seed. As we see, PRL(G), PRL(L), and Co-PRL(L) are robust against different types of corruptions.}
\label{fig:CelebA-a}
\end{figure}

\subsection{Regression Results}

\begin{table*}[t]
	\centering
	\resizebox{1\linewidth}{!}{
\begin{tabular}{|l|l|l|l|l|l|}
\hline
Data                 & $\epsilon-0.1$ & $\epsilon - 0.05$ & $\epsilon$   & $\epsilon + 0.05$ & $\epsilon + 0.1$ \\ \hline
CF10-Pair-45\%       & 65.07±0.83     & 70.07±0.67        & 73.78±0.17   & 77.56±0.55        & 79.36±0.43       \\ \hline
CF10-Sym-50\%        & 69.21±0.35     & 72.53±0.45        & 75.43 ± 0.09 & 77.65±0.27        & 78.10±0.31       \\ \hline
CF10-Sym-70\%        & 53.88±0.64     & 58.49±0.97        & 60.26 ± 0.42 & 60.89±0.43        & 54.91±0.68       \\ \hline
CF100-Pair-45\%      & 32.60±0.45     & 34.17±0.40        & 34.43 ± 0.05 & 36.87±0.41        & 38.34±0.78       \\ \hline
CF100-Sym-50\%       & 37.74±0.41     & 39.72±0.36        & 40.64 ± 0.11 & 43.02±0.36        & 43.92±0.61       \\ \hline
CF100-Sym-70\%       & 24.40±0.47     & 25.50±0.45        & 27.27 ± 0.10 & 27.80±0.50        & 28.20±0.97       \\ \hline
\end{tabular}
}
	\caption{Sensitivity analysis for over-estimated/under-estimated $\epsilon$.}
	\label{tab:sen1}
\end{table*}

For regression, we evaluated our method on the CelebA dataset, which contains 162,770 training images, 19,867 validation images, and 19,962 test images. Given a human face image, the goal is to predict the coordinates for 10 landmarks in the face image. Specifically, the target variable is a ten-dimensional vector of coordinates for the left eye, right eye, nose, left mouth, and right mouth. We added different types of corruption to the landmark coordinates. The CelebA dataset is preprocessed as follows: we use a three-layer CNN to train 162770 training images to predict clean coordinates (we use 19867 validation images to do the early stopping). We then apply the network to extract a 512-dimensional feature vector from the testing data. Thus, the final dataset after preprocessing consists of the feature sets $\rmX \in \mathbb{R}^{19962 \times 512}$ and the target variable $\rmY \in \mathbb{R}^{19962 \times 10}$. We further split the data to the training and testing set, where training sets contain 80\% of the data. We then manually add the \highlight{linadv}, \highlight{signflip}, \highlight{uninoise}, \highlight{pairflip}, and \highlight{mixture} corruptions to the target variable in the training set. The corruption rate for all types of corruptions is varied from 0.1 to 0.4. We use a 3-layer fully connected networks for our experiments. The results of averaged r-square for the last 10 epochs are shown in Table \ref{tab:regression}. The training curves could be found in the figure \ref{fig:CelebA-a}. 
Surprisingly, the performance of PRL(G) is comparable to PRL(L). This is partially due to the network structure and the initialization. Another possible reason is that for this task, the gradient norm is upper bounded by a small constant.

\subsection{Classification Results}
\begin{figure}[]
	\centering
	\begin{subfigure}{0.3\textwidth}
		\includegraphics[width=\linewidth]{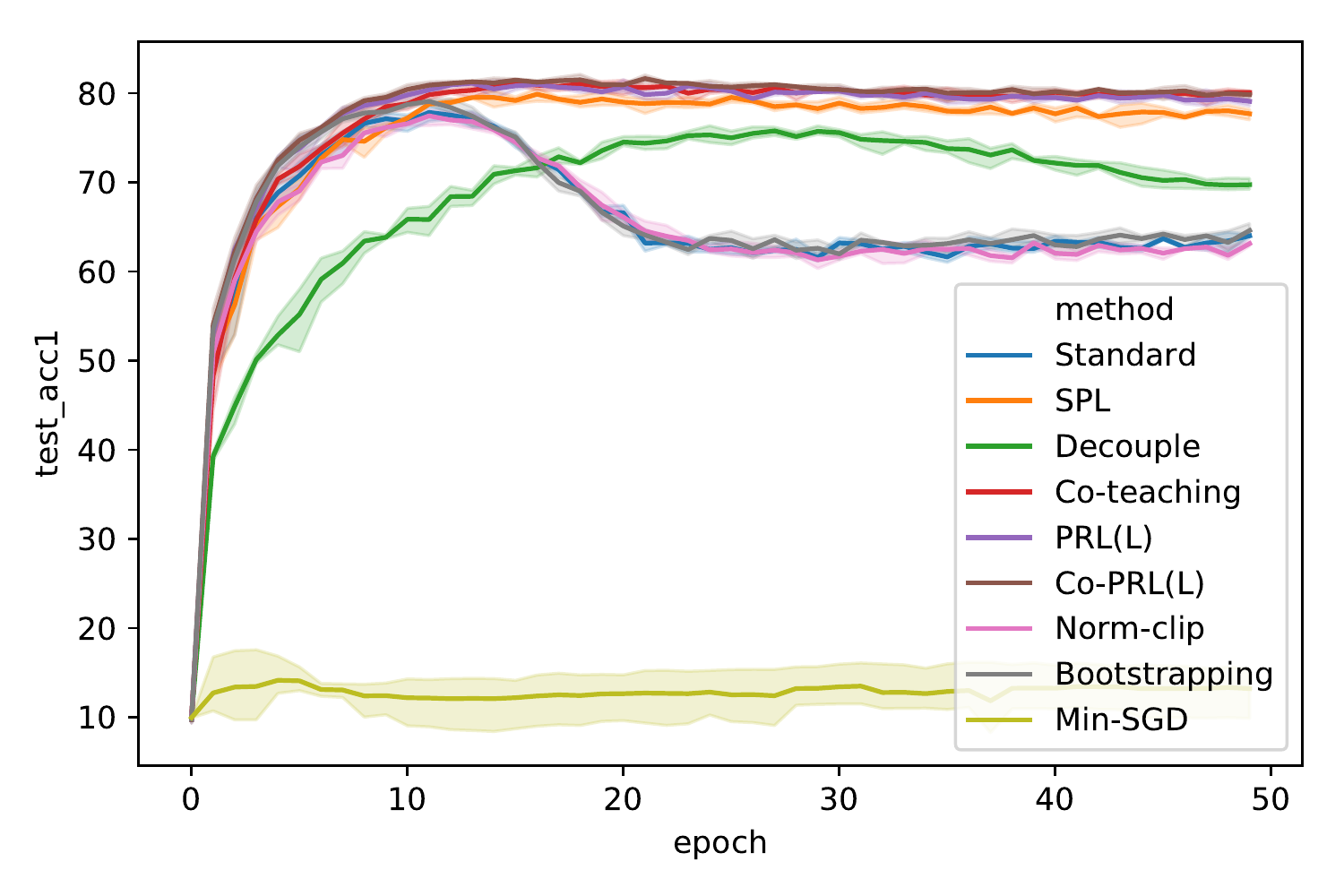}
		\subcaption{CF10 with $\epsilon=0.3$ symmetric noise}
	\end{subfigure}
	\begin{subfigure}{0.3\textwidth}
		\includegraphics[width=\linewidth]{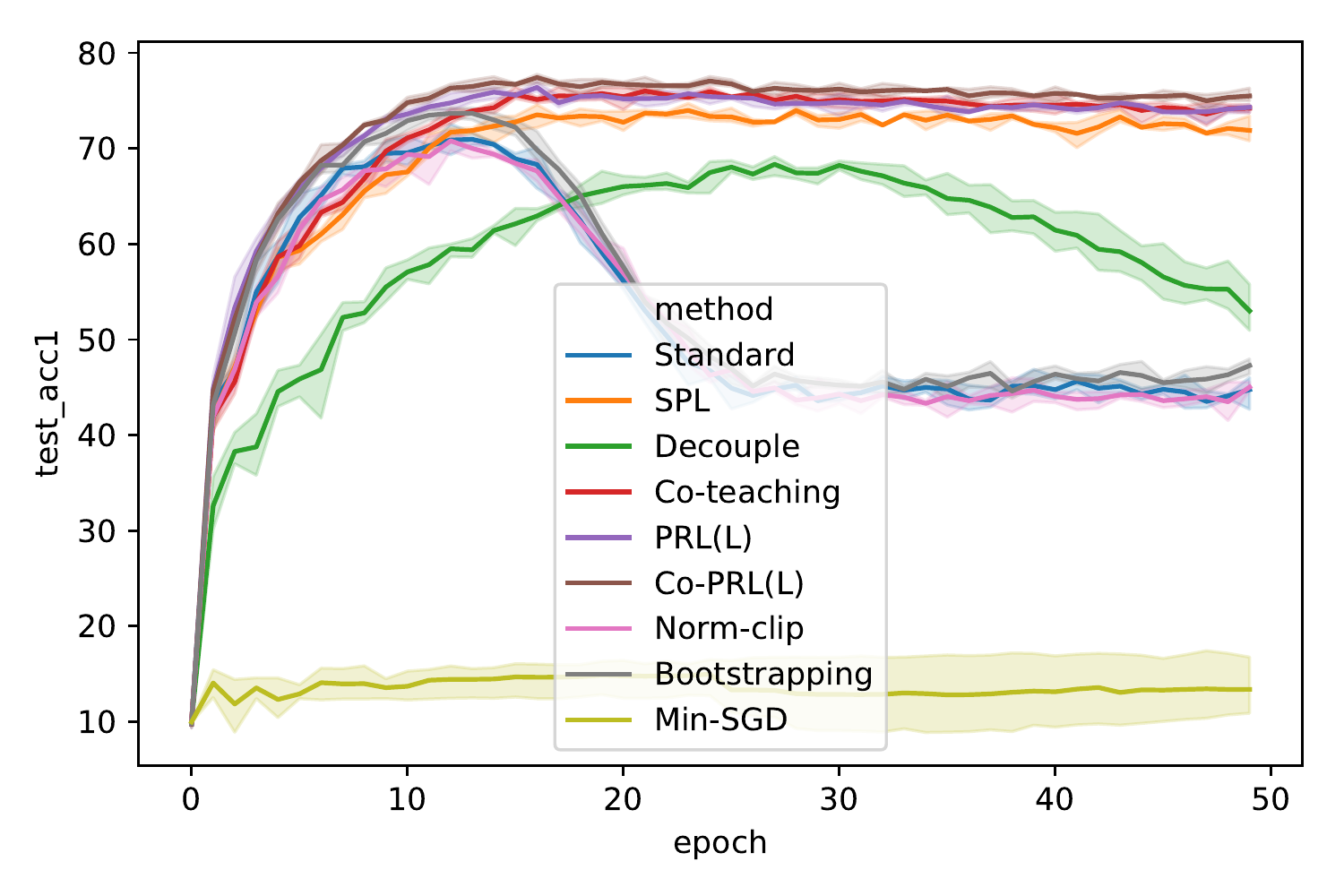}
		\subcaption{CF10 with $\epsilon=0.5$ symmetric noise}
	\end{subfigure}
	\begin{subfigure}{0.3\textwidth}
	\includegraphics[width=\linewidth]{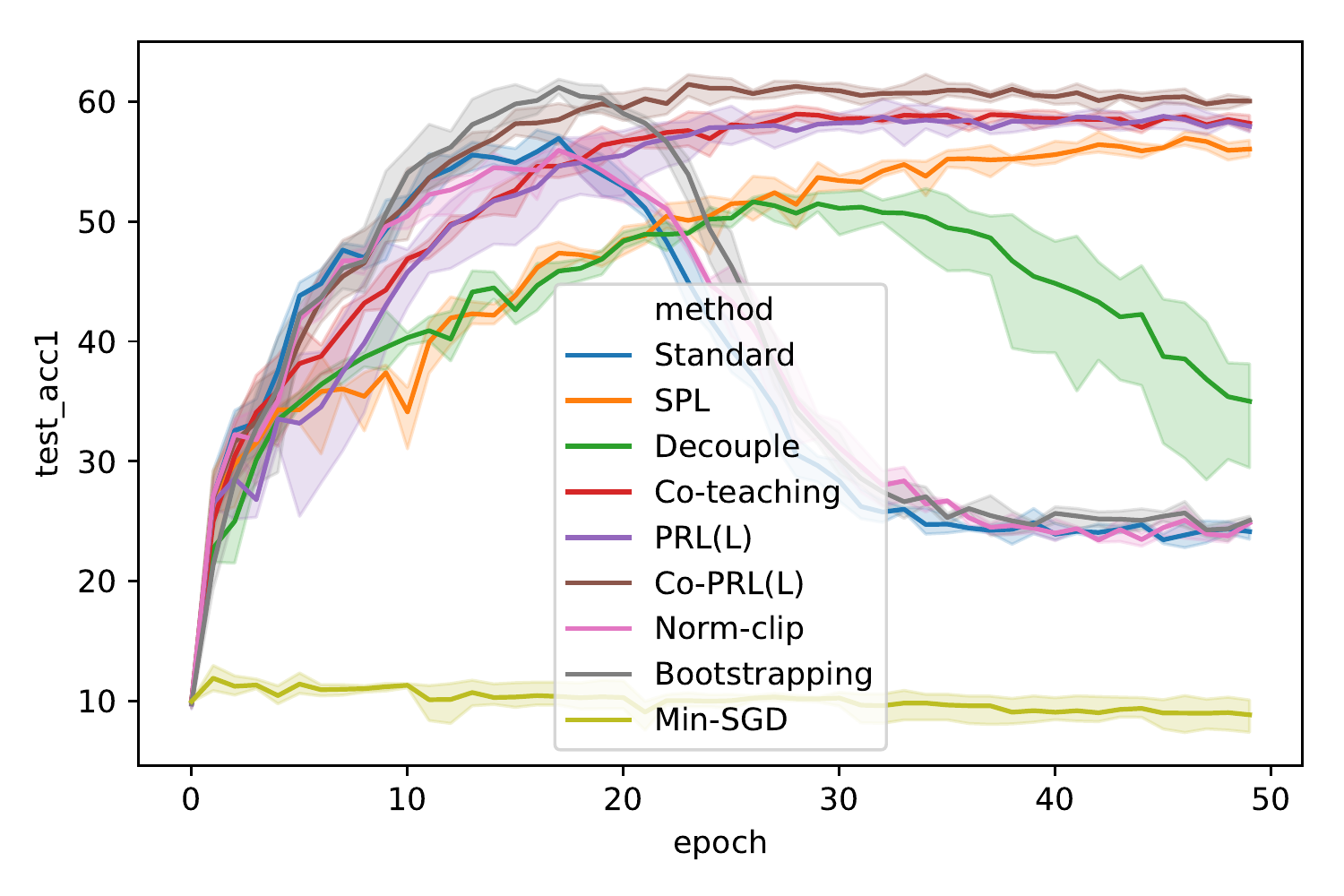}
	\subcaption{CF10 with $\epsilon=0.7$ symmetric noise}
\end{subfigure}
\qquad
    	\begin{subfigure}{0.3\textwidth}
    	\includegraphics[width=\linewidth]{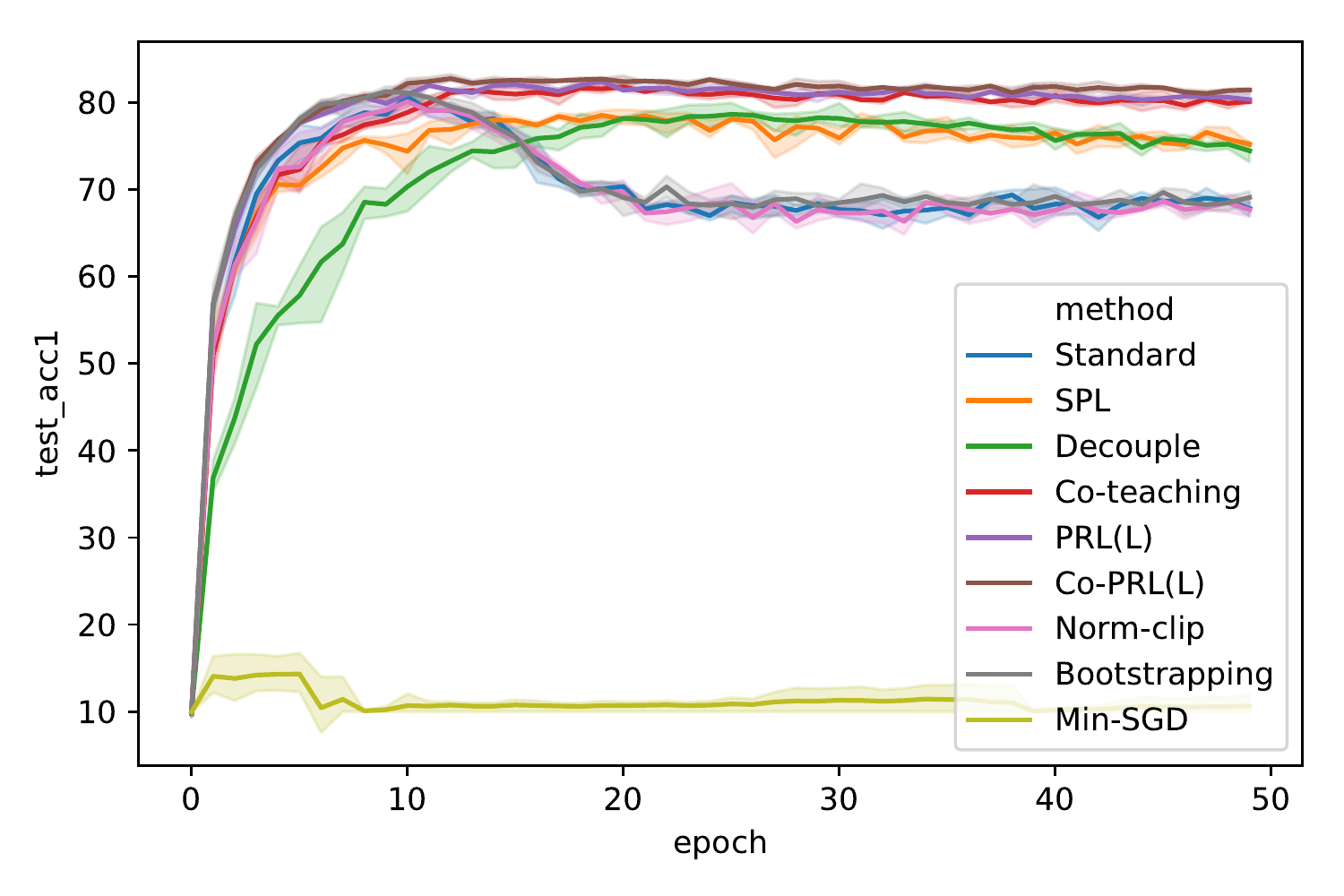}
    	\subcaption{CF10 with $\epsilon=0.25$ pairflip noise}
    \end{subfigure}
    	\begin{subfigure}{0.3\textwidth}
    	\includegraphics[width=\linewidth]{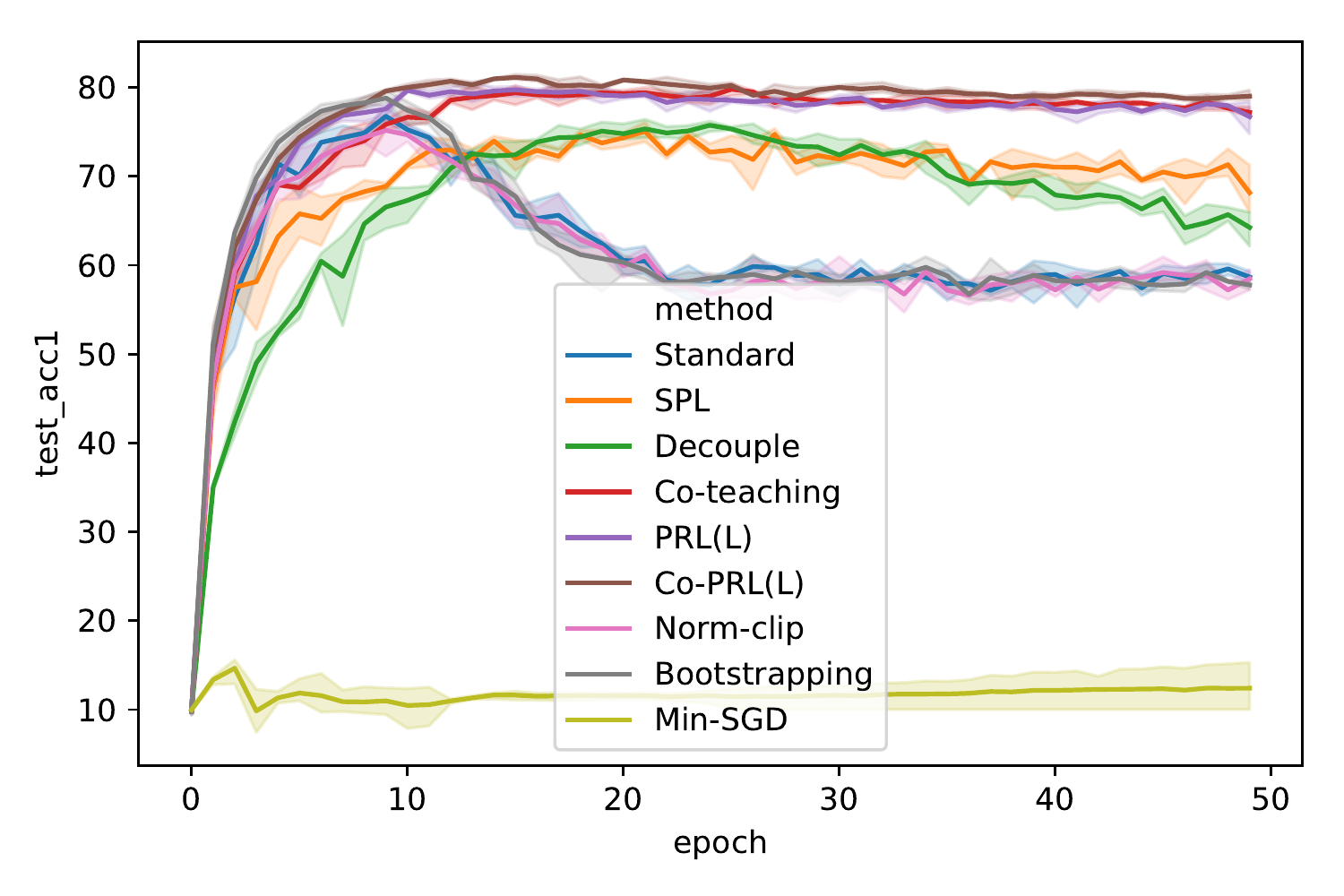}
    	\subcaption{CF10 with $\epsilon=0.35$ pairflip noise}
    \end{subfigure}
    \begin{subfigure}{0.3\textwidth}
    	\includegraphics[width=\linewidth]{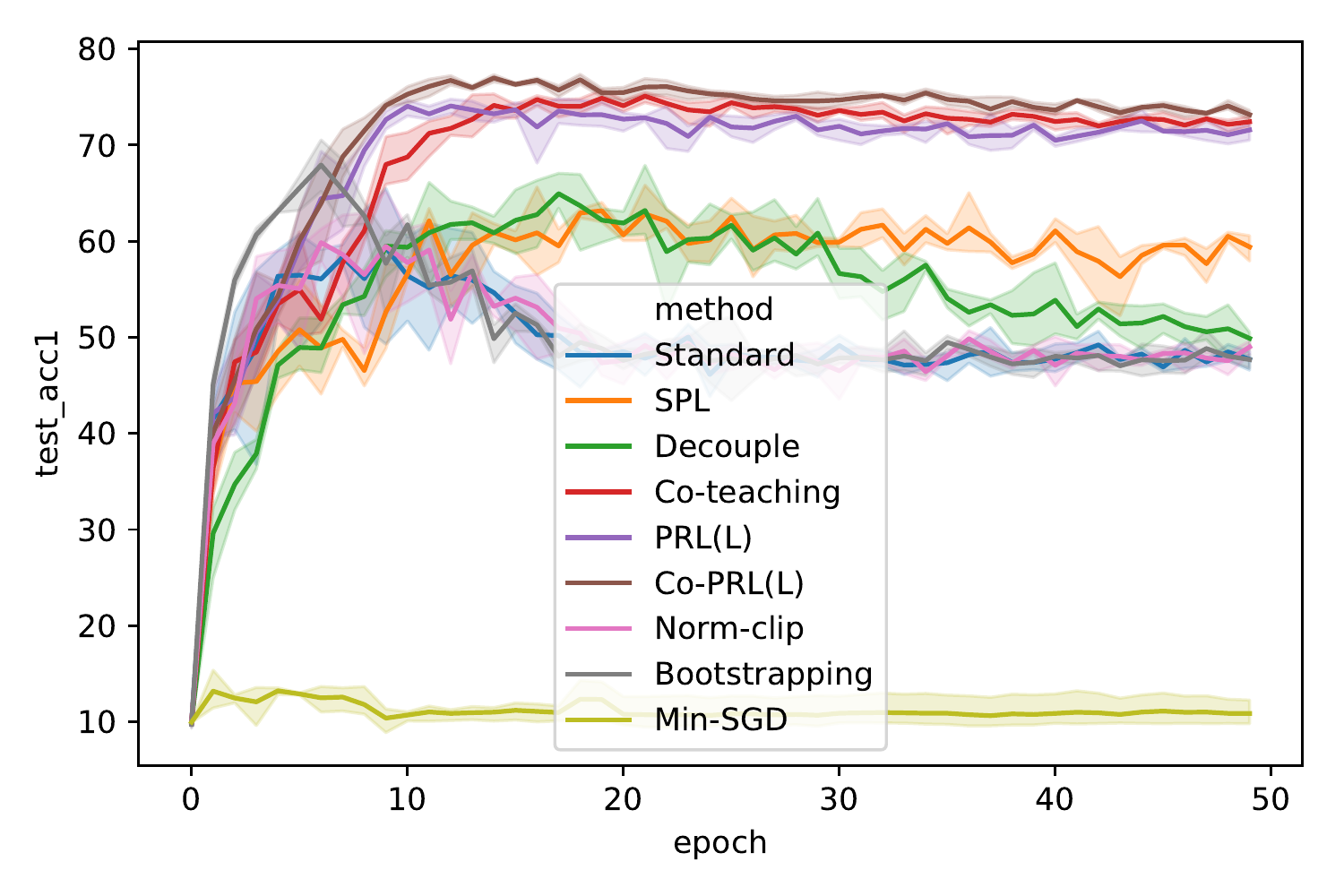}
    	\subcaption{CF10 with $\epsilon=0.45$ pairflip noise}
    \end{subfigure}
\qquad
    \begin{subfigure}{0.3\textwidth}
    	\includegraphics[width=\linewidth]{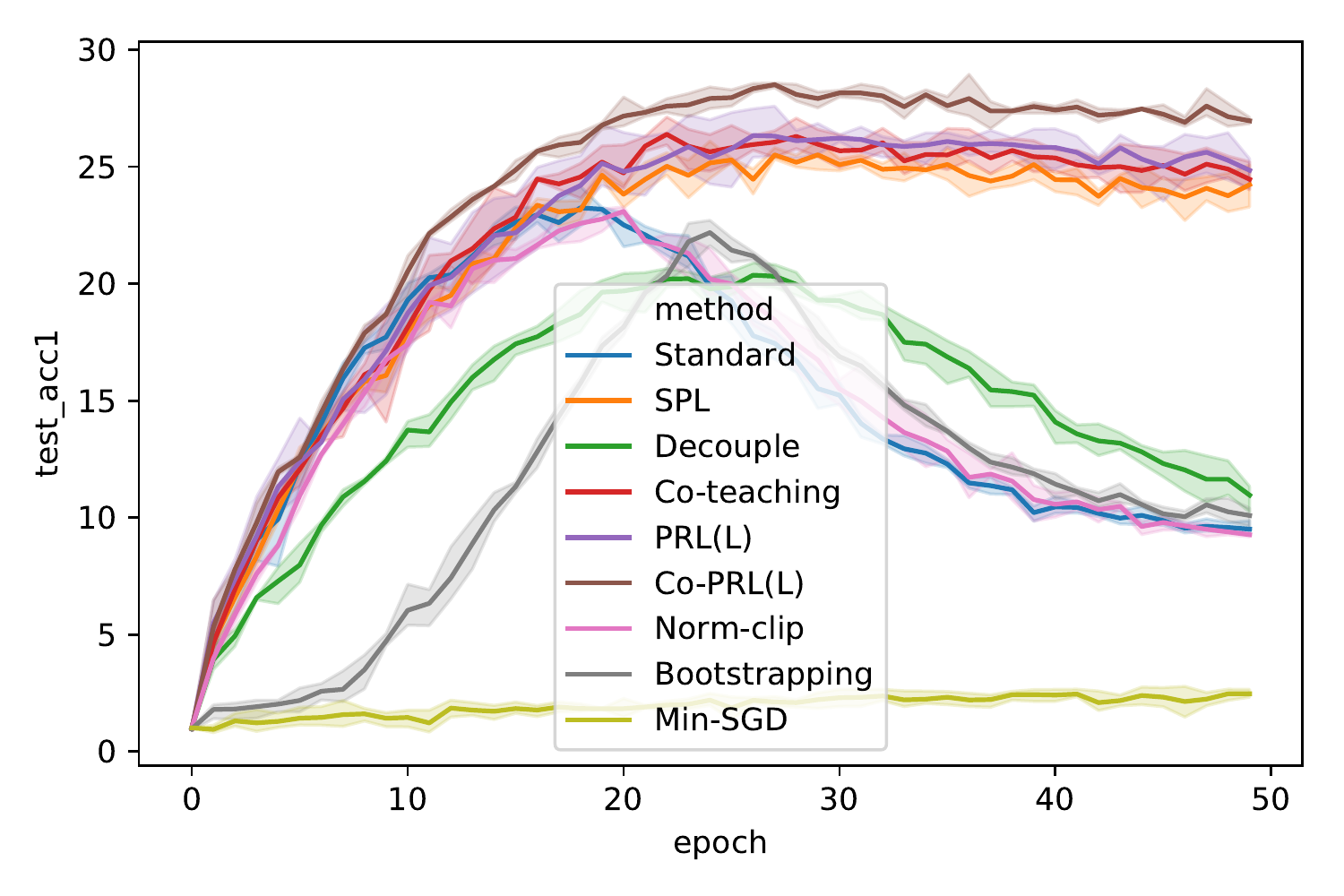}
    	\subcaption{CF100 with $\epsilon=0.3$ symmetric noise}
    \end{subfigure}
    \begin{subfigure}{0.3\textwidth}
    	\includegraphics[width=\linewidth]{Figs/cifar100_symmetric_0.7_acc.pdf}
    	\subcaption{CF100 with $\epsilon=0.5$ symmetric noise}
    \end{subfigure}
    \begin{subfigure}{0.3\textwidth}
    \includegraphics[width=\linewidth]{Figs/cifar100_symmetric_0.7_acc.pdf}
    \subcaption{CF100 with $\epsilon=0.7$ symmetric noise}
    \end{subfigure}
\qquad
    \begin{subfigure}{0.3\textwidth}
    	\includegraphics[width=\linewidth]{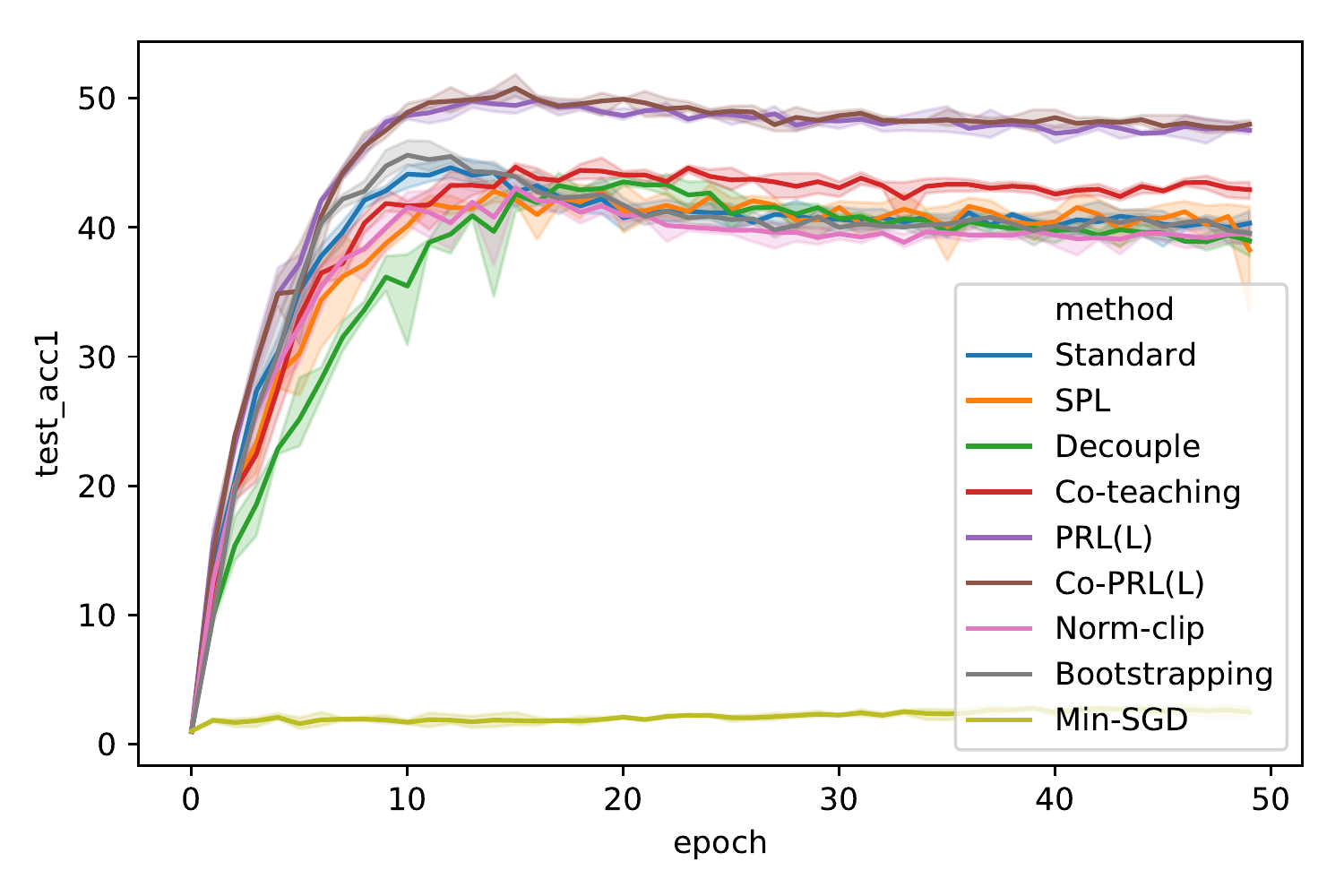}
    	\subcaption{CF100 with $\epsilon=0.25$ pairflip noise}
    \end{subfigure}
    \begin{subfigure}{0.3\textwidth}
    	\includegraphics[width=\linewidth]{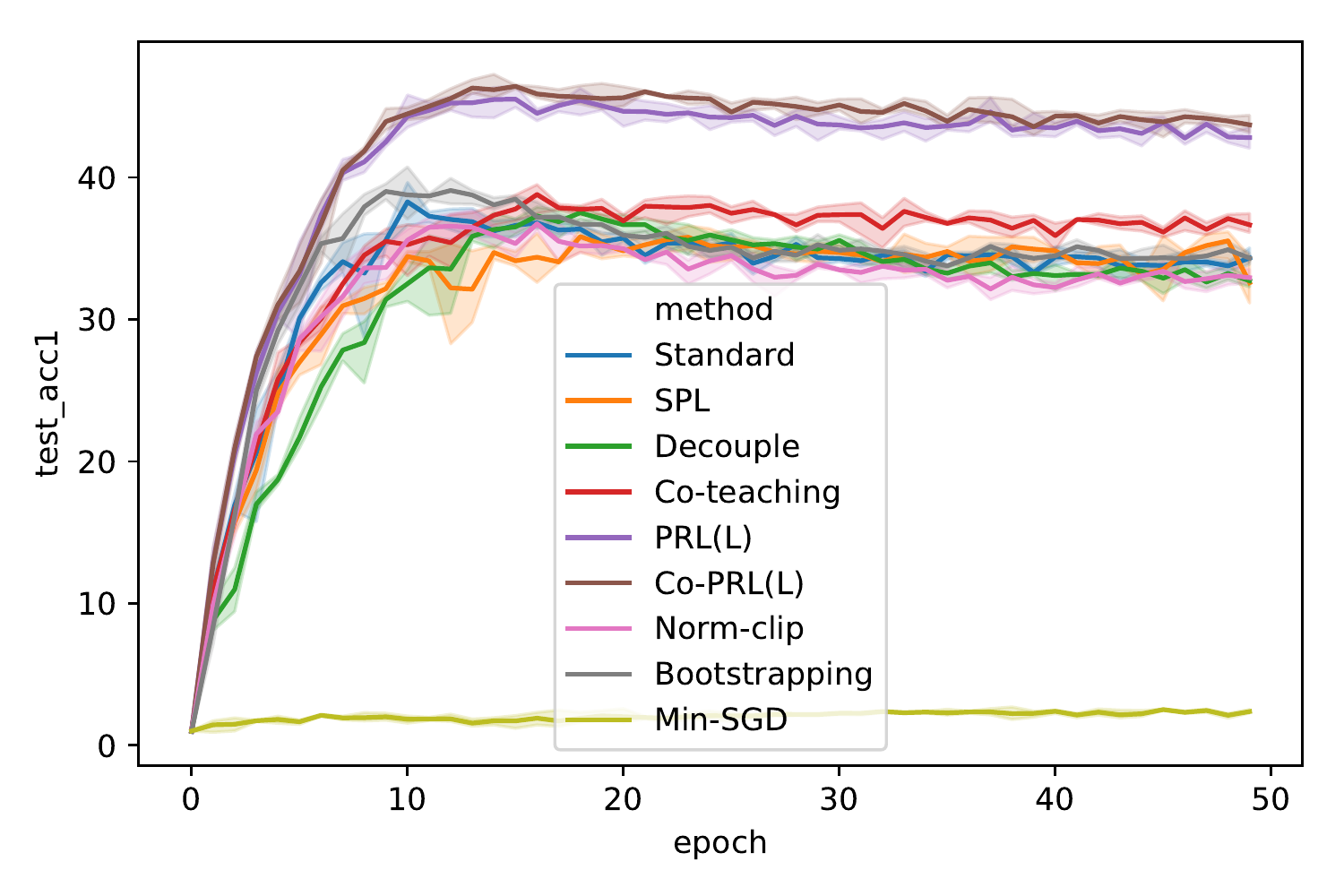}
    	\subcaption{CF100 with $\epsilon=0.35$ pairflip noise}
    \end{subfigure}
    \begin{subfigure}{0.3\textwidth}
    	\includegraphics[width=\linewidth]{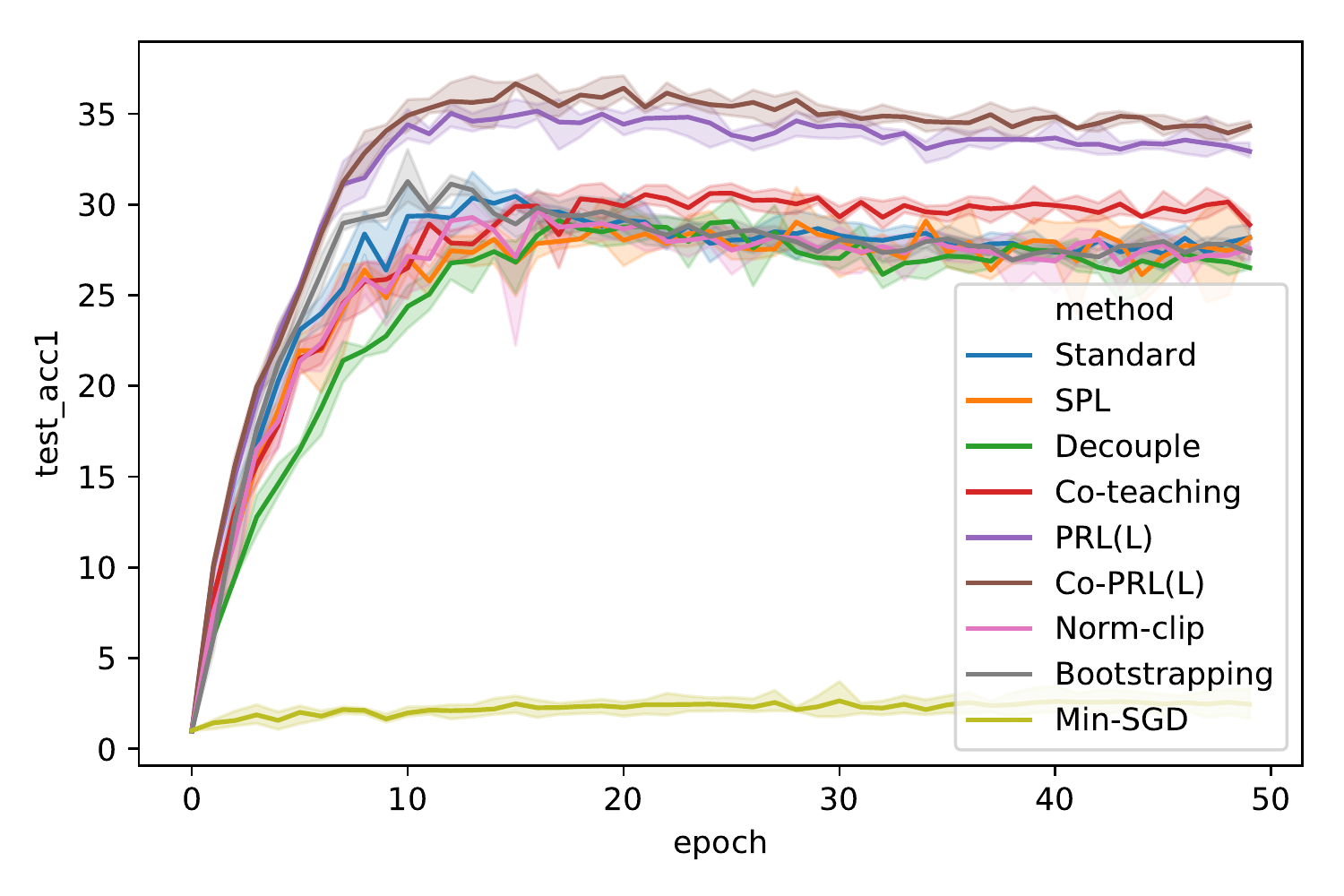}
    	\subcaption{CF100 with $\epsilon=0.45$ pairflip noise}
    \end{subfigure}

\caption{CIFAR10 and CIFAR100 Testing Curve During Training. X axis represents the epoch number, Y axis represents the testing accuracy. The shadow represents the confidence interval, which is calculated across 3 random seed. As we see, PRL(L), and Co-PRL(L) are robust against different types of corruptions.}
\label{fig:CF-curve}
\vspace{1cm}
\end{figure}

We perform our experiments on the CIFAR10 and CIFAR100 datasets to illustrate the effectiveness of our algorithm in classification setting. We use a 9-layer Convolutional Neural Network, similar to the approach in \cite{han2018co}. Since most baselines include batch normalization, it is difficult to get individual gradient efficiently, we exclude the ignormclip and PRL baselines. In the appendix, we attached the results if both co-teaching and Co-PRL(L) excludes the batch normalization module. Our results suggest that co-teaching cannot maintain robustness unlike our proposed method. 
The reason is discussed in the appendix.
We consider \highlight{pairflip} and \highlight{symmetric} supervision corruptions in our experiments. Also, to compare with the current state of the art method, for \highlight{symmetric} noise, we use corruption rate  beyond 0.5. Although our theoretical analysis assumes the noise rate is smaller than 0.5, when the noise type is not an adversary (i.e. symmetric), we empirically show that our method can also deal with such type of noise. The results for CIFAR10 and CIFAR100 are shown in Table \ref{tab:classification}. The results suggest that our method performs significantly better than the baselines irrespective of whether we are using one network (PRL vs SPL) or two networks (Co-PRL(L) vs Co-teaching). The training curves could be found in the figure \ref{fig:CF-curve}. 

\subsection{Sensitivity Analysis}
Since in real-world problems, it is hard to know that the ground-truth corruption rate, we also perform the sensitivity analysis in classification tasks to show the effect of overestimating and underestimating $\epsilon$. The results are in Table \ref{tab:sen1}. As we could see, the performance is stable if we overestimate the corruption rate, this is because only when we overestimate the $\epsilon$, we could guarantee that the gradient norm of the remaining set is small. However, when we underestimate the corruption rate, in the worst case, there is no guarantee that the gradient norm of the remaining set is small. By using the empirical mean, even one large bad individual gradient would ruin the gradient estimation, and according to the convergence analysis of biased gradient descent, the final solution could be very bad in terms of clean data. That explains why to underestimate the corruption rate gives bad results. Also, from Table \ref{tab:sen1}, we could see that using the ground truth corruption rate will lead to small uncertainty.

\section{Conclusion}
In this paper, we proposed a simple yet effective algorithm to defend against agnostic supervision corruptions. Both the theoretical and empirical analysis showed the effectiveness of our algorithm. For future research, there are two 
questions that deserved further study. 
The first question is whether we can further improve $\mathcal{O}(\epsilon)$ error bound or show that $\mathcal{O}(\epsilon)$ is tight. The second question is how we can utilize more properties of neural networks, such as the sparse or low-rank structure in gradient to design better algorithms.

\bibliography{reference}
\bibliographystyle{apalike}

\section{Appendix}
\subsection{Empirical Results on Running Time}
As we claimed in paper, the algorithm 2 (PRL(G)) is not efficient. In here we attached the execution time for one epoch for three different methods: \highlight{Standard}, \highlight{PRL(G)}, \highlight{PRL(L)}. For fair comparison, we replace all batch normalization module to group normalization for this comparison, since it is hard to calculate individual gradient when using batch normalization. For PRL(G), we use opacus libarary (https://opacus.ai/) to calculate the individual gradient.

The results are showed in Table \ref{tab:time}
\begin{table}[]
\centering
\begin{tabular}{|l|l|l|l|}
\hline
Method               & Standard (Lower Bound) &PRL(G) & PRL(L) \\ \hline
CF10-Pair-45\%      & 37.03s        &   145.55s       &    54.80s       \\ \hline
\end{tabular}
\caption{Execution Time of Single Epoch in CIFAR-10 Data}
\label{tab:time}
\end{table}

\subsection{Proof of Theorem \ref{theo:convergence}}
Since this is a standard results, similar results are showed in \cite{bernstein2018signsgd,  devolder2014first, hu2020biased, ajalloeian2020analysis}. For the sake of completeness, we provide the proof sketch here. Details could be found on above literature\\
Proof: 
by L-smooth, we have:
\begin{align*}
\phi(\theta_2) \leq \phi(\theta_1) + \langle \nabla \phi(\theta_1), \theta_2-\theta_1\rangle
 + \dfrac{L}{2}\norm{\theta_2-\theta_1}^2    
\end{align*}
by using $\gamma \leq \dfrac{1}{L}$, we have
\begin{align*}
\mathbb{E} \phi\left(\mathbf{\theta_1}_{t+1}\right) & \leq \phi\left(\mathbf{\theta_1}_{t}\right)-\gamma\left\langle\nabla \phi\left(\mathbf{\theta_1}_{t}\right), \mathbb{E} \mathbf{g}_{t}\right\rangle+\frac{\gamma^{2} L}{2}\left(\mathbb{E}\left\|\mathbf{g}_{t}-\mathbb{E} \mathbf{g}_{t}\right\|^{2}+\mathbb{E}\left\|\mathbb{E} \mathbf{g}_{t}\right\|^{2}\right) \\
&=\phi\left(\mathbf{\theta_1}_{t}\right)-\gamma\left\langle\nabla \phi\left(\mathbf{\theta_1}_{t}\right), \nabla \phi\left(\mathbf{\theta_1}_{t}\right)+\mathbf{b}_{t}\right\rangle+\frac{\gamma^{2} L}{2}\left(\mathbb{E}\left\|\mathbf{n}_{t}\right\|^{2}+\mathbb{E}\left\|\nabla \phi\left(\mathbf{\theta_1}_{t}\right)+\mathbf{b}_{t}\right\|^{2}\right) \\
& \leq \phi\left(\mathbf{\theta_1}_{t}\right)+\frac{\gamma}{2}\left(-2\left\langle\nabla \phi\left(\mathbf{\theta_1}_{t}\right), \nabla \phi\left(\mathbf{\theta_1}_{t}\right)+\mathbf{b}_{t}\right\rangle+\left\|\nabla \phi\left(\mathbf{\theta_1}_{t}\right)+\mathbf{b}_{t}\right\|^{2}\right)+\frac{\gamma^{2} L}{2} \mathbb{E}\left\|\mathbf{n}_{t}\right\|^{2} \\
&=\phi\left(\mathbf{\theta_1}_{t}\right)+\frac{\gamma}{2}\left(-\left\|\nabla \phi\left(\mathbf{\theta_1}_{t}\right)\right\|^{2}+\left\|\mathbf{b}_{t}\right\|^{2}\right)+\frac{\gamma^{2} L}{2} \mathbb{E}\left\|\mathbf{n}_{t}\right\|^{2}
\end{align*}
According to the assumption, we have $\norm{\mathbf{b}_{t}}^2 \leq \zeta^{2}$, $\norm{\mathbf{n}_{t}}^2 \leq \sigma^{2}$, by plug in the learning rate constraint, we have
\begin{align*}
    &\mathbb{E} \phi\left(\mathbf{\theta_1}_{t+1}\right) \leq \phi\left(\mathbf{\theta_1}_{t}\right)-\frac{\gamma}{2}\left\|\nabla \phi\left(\mathbf{\theta_1}_{t}\right)\right\|^{2}+\frac{\gamma}{2} \zeta^{2}+\frac{\gamma^{2} L}{2} \sigma^{2}\\
	&\mathbb{E} \phi\left(\mathbf{\theta_1}_{t+1}\right)-\phi\left(\mathbf{\theta_1}_{t}\right)
	\leq -\frac{\gamma}{2}\left\|\nabla \phi\left(\mathbf{\theta_1}_{t}\right)\right\|^{2}+\frac{\gamma}{2} \zeta^{2}+\frac{\gamma^{2} L}{2} \sigma^{2}
\end{align*}
Then, removing the gradient norm to left hand side, and sum it across different iterations, we could get 
\begin{align*}
    \dfrac{1}{2T}\sum_{t=0}^{T-1}\mathbb{E}\norm{\phi\left(\mathbf{\theta_1}_{t}\right)} &\leq \frac{F}{T \gamma}+\frac{\zeta^{2}}{2}+\frac{\gamma L \sigma^{2}}{2}\\
\end{align*}
Take the minimum respect to t and substitute the learning rate condition will directly get the results.

\subsection{Proof of Corollary \ref{coro:ro}}
We first prove the gradient estimation error.

Denote $\tilde{\rmG}$ to be the set of corrupted mini-batch,  $\rmG$ to be the set of original clean mini-batch and we have $|\rmG| = |\tilde{\rmG}|  = m$. Let $\rmN$ to be the set of remaining data and according to our algorithm, the remaining data has the size $|\rmN| = n = (1-\epsilon)m$. Define $\rmA$ to be the set of individual clean gradient, which is not discarded by algorithm 1. $\rmB$ to be the set of individual corrupted gradient, which is not discarded. According to our definition, we have $\rmN = \rmA \cup \rmB$. $\mathbf{AD}$ to be the set of individual good gradient, which is discarded, $\mathbf{AR}$ to be the set of individual good gradient, which is replaced by corrupted data. We have $\rmG = \rmA\cup \mathbf{AD} \cup \mathbf{AR}$. $\mathbf{BD}$ is  the set of individual corrupted gradient, which is discarded by our algorithm. Denote the good gradient to be $\rvg_i = \alpha_i \rmW_i$, and the bad gradient to be $\tilde{\rvg}_i$,  according to our assumption, we have $\norm{\tilde{\rvg}_i} \leq L$.

Now, we have the l2 norm error:
\begin{align}
\norm{\mu(\rmG) - \mu(\rmN)} &= \norm{\dfrac{1}{m}\sum_{i \in \rmG}^{m}\rvg_i - \left(\dfrac{1}{n}\sum_{i \in \rmA} \rvg_i + \dfrac{1}{n}\sum_{i \in \rmB}\tilde{\rvg}_i\right)} \nonumber\\
&=\norm{\dfrac{1}{n}\sum_{i=1}^m \dfrac{n}{m}\rvg_i - \left(\dfrac{1}{n}\sum_{i \in \rmA} \rvg_i + \dfrac{1}{n}\sum_{i \in \rmB} \tilde{\rvg}_i\right)} \nonumber\\
&=\norm{\dfrac{1}{n}\sum_{i \in \rmA}\dfrac{n}{m} \rvg_i + \dfrac{1}{n}\sum_{i \in \mathbf{ AD}}\dfrac{n}{m} \rvg_i + \dfrac{1}{n}\sum_{i \in \mathbf{AR}}\dfrac{n}{m} \rvg_i - \left(\dfrac{1}{n}\sum_{i \in \rmA} \rvg_i + \dfrac{1}{n}\sum_{i \in \rmB}\tilde{\rvg}_i\right)} \nonumber\\
&=\norm{\dfrac{1}{n}\sum_{i \in \rmA}(\dfrac{n-m}{m})\rvg_i + \dfrac{1}{n}\sum_{i \in \mathbf{AD}}\dfrac{n}{m}\rvg_i + \dfrac{1}{n}\sum_{i \in \mathbf{AR}}\dfrac{n}{m} \rvg_i- \dfrac{1}{n}\sum_{i \in \mathbf{B}} \tilde{\rvg}_i} \nonumber\\
&\leq \norm{\dfrac{1}{n}\sum_{i \in \rmA}(\dfrac{n-m}{m})\rvg_i + \dfrac{1}{n}\sum_{i \in\mathbf{AD}}\dfrac{n}{m}\rvg_i + \dfrac{1}{n}\sum_{i \in \mathbf{AR}}\dfrac{n}{m} \rvg_i} +\norm{\dfrac{1}{n}\sum_{i \in \rmB} \tilde{\rvg}_i} \nonumber\\
&\leq  \norm{\sum_{\rmA}\dfrac{m-n}{nm}\rvg_i + \sum_{\mathbf{AD}}\dfrac{1}{m}\rvg_i + \sum_{\mathbf{AR}}\dfrac{1}{m}\rvg_i}  + \sum_{\mathbf{B}}\dfrac{1}{n}\norm{\tilde{\rvg}_i} \nonumber\\
&\leq \sum_{\rmA}\norm{\dfrac{m-n}{nm}\rvg_i} + \sum_{\mathbf{AD}}\norm{\dfrac{1}{m}\rvg_i} + \sum_{\mathbf{AR}}\norm{\dfrac{1}{m}\rvg_i} + \sum_{\mathbf{\rmB}}\dfrac{1}{n}\norm{\tilde{\rvg}_i} \nonumber
\end{align}

By using the filtering algorithm, we could guarantee that $\norm{\tilde{\rvg}_i} \leq L$.
Let $|\rmA| = x$, we have $|\rmB| = n - x = (1-\epsilon)m - x$, $|\mathbf{AR}| = m - n = \epsilon m$, $|\mathbf{AD}| = m - |\rmA| - |\mathbf{AR}| = m - x - (m - n) = n-x = (1-\epsilon)m - x$.
Thus, we have:
\begin{align*}
\norm{\mu(\rmG) - \mu(\rmN)} &\leq  x\dfrac{m-n}{nm}L + (n-x)\dfrac{1}{m}L + (m-n)\dfrac{1}{m}L + (n-x)\dfrac{1}{n}L \\
&\leq x(\dfrac{m-n}{nm} - \dfrac{1}{m})L + n\dfrac{1}{m}L + (m-n)\dfrac{1}{m} L + (n-x)\dfrac{1}{n}L\\
&=\dfrac{1}{m}(\dfrac{2\epsilon-1}{1-\epsilon})xL + L + L - \dfrac{1}{n}xL\\
&=xL(\dfrac{2\epsilon-2}{n}) + 2L
\end{align*}
To minimize the upper bound, we need $x$ to be as small as possible since $2\epsilon-2 < 1$. According to our problem setting, we have $x = n - m\epsilon \leq (1-2\epsilon)m$, substitute back we have:
\begin{align*}
\norm{\mu(\rmG) - \mu(\rmN)} &\leq  (1-2\epsilon)Lm(\dfrac{2\epsilon-2}{n}) + 2L\\
& = \dfrac{1-2\epsilon}{1-\epsilon}2L + 2L\\
&=4L - \dfrac{\epsilon}{1-\epsilon}2L
\end{align*}
Since $\epsilon < 0.5$, we use tylor expansion on $\dfrac{\epsilon}{1-\epsilon}$, by ignoring the high-order terms, we have
\begin{align*}
\norm{\mu(\rmG) - \mu(\rmN)} = \mathcal{O}(\epsilon L)
\end{align*}

Note, if the Lipschitz continuous assumption does not hold, then L should be dimension dependent (i.e. $\sqrt{d}$).

Combining above gradient estimation error upper bound and Theorem 1, we could get the results in Corollary 1.

\subsection{Proof of Lemma \ref{lemma:randomdroperr}}
Denote $\tilde{\rmG}$ to be the set of corrupted mini-batch,  $\rmG$ to be the set of original clean mini-batch and we have $|\rmG| = |\tilde{\rmG}|  = m$. Let $\rmN$ to be the set of remaining data and according to our algorithm, the remaining data has the size $|\rmN| = n = (1-\epsilon)m$. Define $\rmA$ to be the set of individual clean gradient, which is not discarded by any filtering algorithm. $\rmB$ to be the set of individual corrupted gradient, which is not discarded. According to our definition, we have $\rmN = \rmA \cup \rmB$. $\mathbf{AD}$ to be the set of individual good gradient, which is discarded, $\mathbf{AR}$ to be the set of individual good gradient, which is replaced by corrupted data. We have $\rmG = \rmA\cup \mathbf{AD} \cup \mathbf{AR}$. $\mathbf{BD}$ is  the set of individual corrupted gradient, which is discarded by our algorithm. Denote the good gradient to be $\rvg_i = \alpha_i \rmW_i$, and the bad gradient to be $\tilde{\rvg}_i = \delta_i \rmW_i$,  according to our assumption, we have $\norm{\rmW_i}_{op} \leq C$.

Now, we have the l2 norm error:
\begin{align}
\norm{\mu(\rmG) - \mu(\rmN)} &= \norm{\dfrac{1}{m}\sum_{i \in \rmG}^{m}\rvg_i - \left(\dfrac{1}{n}\sum_{i \in \rmA} \rvg_i + \dfrac{1}{n}\sum_{i \in \rmB}\tilde{\rvg}_i\right)} \nonumber\\
&=\norm{\dfrac{1}{n}\sum_{i=1}^m \dfrac{n}{m}\rvg_i - \left(\dfrac{1}{n}\sum_{i \in \rmA} \rvg_i + \dfrac{1}{n}\sum_{i \in \rmB} \tilde{\rvg}_i\right)} \nonumber\\
&=\norm{\dfrac{1}{n}\sum_{i \in \rmA}\dfrac{n}{m} \rvg_i + \dfrac{1}{n}\sum_{i \in \mathbf{ AD}}\dfrac{n}{m} \rvg_i + \dfrac{1}{n}\sum_{i \in \mathbf{AR}}\dfrac{n}{m} \rvg_i - \left(\dfrac{1}{n}\sum_{i \in \rmA} \rvg_i + \dfrac{1}{n}\sum_{i \in \rmB}\tilde{\rvg}_i\right)} \nonumber\\
&=\norm{\dfrac{1}{n}\sum_{i \in \rmA}(\dfrac{n-m}{m})\rvg_i + \dfrac{1}{n}\sum_{i \in \mathbf{AD}}\dfrac{n}{m}\rvg_i + \dfrac{1}{n}\sum_{i \in \mathbf{AR}}\dfrac{n}{m} \rvg_i- \dfrac{1}{n}\sum_{i \in \mathbf{B}} \tilde{\rvg}_i} \nonumber\\
&\leq \norm{\dfrac{1}{n}\sum_{i \in \rmA}(\dfrac{n-m}{m})\rvg_i + \dfrac{1}{n}\sum_{i \in\mathbf{AD}}\dfrac{n}{m}\rvg_i + \dfrac{1}{n}\sum_{i \in \mathbf{AR}}\dfrac{n}{m} \rvg_i} +\norm{\dfrac{1}{n}\sum_{i \in \rmB} \tilde{\rvg}_i} \label{eq:3term}
\end{align}

Let $|\rmA| = x$, we have $|\rmB| = n - x = (1-\epsilon)m - x$, $|\mathbf{AR}| = m - n = \epsilon m$, $|\mathbf{AD}| = m - |\rmA| - |\mathbf{AR}| = m - x - (m - n) = n-x = (1-\epsilon)m - x$.
Thus, we have:
\begin{align*}
\norm{\mu(\rmG) - \mu(\rmN)} &\leq  \norm{\sum_{\rmA}\dfrac{m-n}{nm}\rvg_i + \sum_{\mathbf{AD}}\dfrac{1}{m}\rvg_i + \sum_{\mathbf{AR}}\dfrac{1}{m}\rvg_i}  + \sum_{\mathbf{B}}\dfrac{1}{n}\norm{\tilde{\rvg}_i}\\
&\leq \sum_{\rmA}\norm{\dfrac{m-n}{nm}\rvg_i} + \sum_{\mathbf{AD}}\norm{\dfrac{1}{m}\rvg_i} + \sum_{\mathbf{AR}}\norm{\dfrac{1}{m}\rvg_i} + \sum_{\mathbf{\rmB}}\dfrac{1}{n}\norm{\tilde{\rvg}_i}
\end{align*}
For individual gradient, according to the label corruption gradient definition in problem 2, assuming the $\norm{\rmW}_{op}\leq C$, we have $\norm{\rvg_i} \leq \norm{\alpha_i}\norm{\rmW_i}_{op} \leq C \norm{\alpha_i}$. Also, denote $\max_{i}\norm{\alpha_i} = k$, $\max_{i}\norm{\delta_i} = v$, we have  $\norm{\rvg_i} \leq Ck$, $\norm{\tilde{\rvg_i}} \leq Cv$.
\begin{align*}
\norm{\mu(\rmG) - \mu(\rmN)} &\leq  Cx\dfrac{m-n}{nm} k + C(n-x)\dfrac{1}{m}k + C(m-n)\dfrac{1}{m} k + C(n-x)\dfrac{1}{n}v
\end{align*}
Note the above upper bound holds for any $x$, thus, we would like to get the minimum of the upper bound respect to $x$. Rearrange the term, we have 
\begin{align*}
\norm{\mu(\rmG) - \mu(\rmN)} &\leq  Cx(\dfrac{m-n}{nm} - \dfrac{1}{m}) k + Cn\dfrac{1}{m}k + C(m-n)\dfrac{1}{m} k + C(n-x)\dfrac{1}{n}v\\
&=C\dfrac{1}{m}(\dfrac{2\epsilon-1}{1-\epsilon})xk + Ck + Cv - \dfrac{1}{n}Cxv\\
&=Cx\left(\dfrac{k(2\epsilon-1)}{m(1-\epsilon)} - \dfrac{v}{n}\right) + Ck + Cv\\
&=Cx\left(\dfrac{k(2\epsilon-1)-v}{m(1-\epsilon)}\right) + Ck + Cv
\end{align*}
Since when $\epsilon < 0.5$, $\dfrac{k(2\epsilon-1)-v}{m(1-\epsilon)} < 0$, we knew that $x$ should be as small as possible to continue the bound. According to our algorithm, we knew $n - m\epsilon = m(1-\epsilon)-m\epsilon = (1-2\epsilon)m \leq x \leq n = (1-\epsilon)m$. Then, substitute $x = (1-2\epsilon)m$, we have
\begin{align*}
\norm{\mu(\rmG) - \mu(\rmN)} &\leq Ck(1-2\epsilon)\dfrac{2\epsilon-1}{1-\epsilon} + Ck + Cv - Cv\dfrac{1-2\epsilon}{1-\epsilon}\\
&=Ck\dfrac{3\epsilon - 4\epsilon^2}{1-\epsilon} + Cv\dfrac{\epsilon}{1-\epsilon}
\end{align*}

\subsection{Proof of Theorem \ref{theo:mlrme}}
According to algorithm2, we could guarantee that $v \leq k$. By lemma 1, we will have:
\begin{align*}
\norm{\mu(\rmG) - \mu(\rmN)} &\leq Ck\dfrac{3\epsilon - 4\epsilon^2}{1-\epsilon} + Cv\dfrac{\epsilon}{1-\epsilon}\\
&\leq Ck\dfrac{4\epsilon - 4\epsilon^2}{1-\epsilon}\\
&=4\epsilon Ck\\
&\approx\mathcal{O}(\epsilon \sqrt{q}) \text{(C is constant, k is the norm of $q$-dimensional vector)}
\end{align*}

\subsection{Proof of Lemma 2}
Assume we have a $d$ class label $\mathbf{y} \in \mathcal{R}^d$, where $y_k=1,y_i=0,i \neq k$. We have two prediction $\mathbf{p} \in \mathcal{R}^d$, $\mathbf{q} \in \mathcal{R}^d$.

Assume we have a $d$ class label $\mathbf{y} \in \mathbb{R}^d$, where $y_k=1,y_i=0,i \neq k$. With little abuse of notation, suppose we have two prediction $\mathbf{p} \in \mathbb{R}^d$, $\mathbf{q} \in \mathbb{R}^d$. Without loss of generality, we could assume that $\rvp_1$ has smaller cross entropy loss, which indicates $\rvp_k \geq \rvq_k$

For MSE, assume we have opposite result
\begin{equation}
\begin{split}
&\Vert \mathbf{p}-\mathbf{y}\Vert^2 \geq \Vert \mathbf{q}-\mathbf{y}\Vert^2 \\
\Rightarrow &\sum_{i \neq k} p_i^2 + (1-p_k)^2 \geq \sum_{i \neq k} q_i^2 + (1-q_k)^2
\end{split}
\end{equation}

For each $p_i, i \neq k$, We have
\begin{equation}
\begin{split}
Var(p_i) &=  E(p_i^2) - E(p_i)^2 = \frac{1}{d-1}\sum_{i \neq k}{p_i^2}-\frac{1}{(d-1)^2}(1-p_k)^2
\end{split}
\end{equation}

Then
\begin{equation}
\begin{split}
&\sum_{i \neq k} p_i^2 + (1-p_k)^2 \geq \sum_{i \neq k} q_i^2 + (1-q_k)^2\\
\Rightarrow & Var_{i \neq k}(\rvp_i) + \frac{d}{(d-1)^2}(1-p_k)^2 \geq Var_{i \neq k}(\rvq_i) + \frac{d}{(d-1)^2}(1-q_k)^2\\
\Rightarrow & Var_{i \neq k}(\rvp_i) - Var_{i \neq k}(\rvq_i)  \geq \frac{d}{(d-1)^2}\left((1-q_k)^2 -(1-p_k)^2\right) \\
\Rightarrow & Var_{i \neq k}(\rvp_i) - Var_{i \neq k}(\rvq_i)  \geq \frac{d}{(d-1)^2}\left((p_k-q_k)(2-p_k-q_k)\right)
\end{split}
\end{equation}

\end{document}